\documentclass[journal]{IEEEtran}

\usepackage{amsmath}
\usepackage{amssymb}
\usepackage{booktabs}
\usepackage{cite}
\usepackage{floatrow}
\usepackage{graphicx}
\usepackage{hyperref}
\usepackage{multirow}
\usepackage{setspace}
\usepackage{tabularx}
\usepackage{threeparttable}
\usepackage[caption=false,font=footnotesize]{subfig}
\usepackage[x11names]{xcolor}

\newcommand{\ie}{\textit{i.e.,} }
\newcommand{\eg}{\textit{e.g.,} }
\newcommand{\etal}{\textit{et al.} }

\newcommand{\orcid}[1]{\hspace{0.5mm}\raisebox{0.85mm}{\href{https://orcid.org/#1}{\includegraphics[width=8pt]{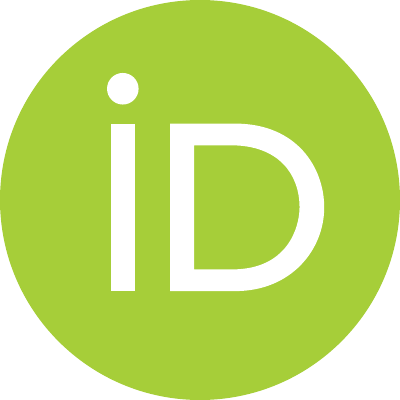}}}\hspace{0.2mm}}
\newcommand{\email}[1]{\href{mailto:#1}{#1}}
\newcommand{\oparallel}{\protect\ooalign{$\vcenter{\hbox{\scalebox{0.75}{$\bigcirc$}}}$\cr\hidewidth$\vcenter{\hbox{\scalebox{0.65}{$\parallel$}}}$\hidewidth\cr}}
\newcommand{\orelu}{\protect\ooalign{$\vcenter{\hbox{\scalebox{0.75}{$\bigcirc$}}}$\cr$\vbox{\hbox{\hspace{0.2pt}\vspace{-0.6pt}\scalebox{0.6}{$-$}}}$\cr$\vbox{\hbox{\hspace{2pt}\vspace{-0.15pt}\rotatebox{60}{\scalebox{0.6}{$-$}}}}$}}

\begin{document}

\title{Learning to Aggregate Multi-Scale Context for Instance Segmentation in Remote Sensing Images}

\author{
Ye~Liu\orcid{0000-0001-9597-0525},
Huifang~Li\orcid{0000-0003-4626-7416},~\IEEEmembership{Member,~IEEE},
Chao~Hu\orcid{0000-0001-6183-9051},
Shuang~Luo\orcid{0000-0003-3832-0475},
Yan~Luo\orcid{0000-0002-9533-6070},\\
and~Chang~Wen~Chen\orcid{0000-0002-6720-234X},~\IEEEmembership{Fellow,~IEEE}
\thanks{\\\indent Manuscript received March 22, 2022; revised June 19, 2023 and August 24, 2023; accepted November 20, 2023. This work was supported by the National Natural Science Foundation of China under Grant 42371366, the Hubei Provincial Key Research and Development Program under Grant 2023BAB066, and the Fundamental Research Funds for the Central Universities under Grant 2042023kfyq04. \textit{(Corresponding author: Huifang Li)}}
\thanks{Ye Liu is with the Department of Computing, The Hong Kong Polytechnic University, Hong Kong SAR, China (e-mail: \email{coco.ye.liu@connect.polyu.hk}).}
\thanks{Huifang Li and Chao Hu are with the School of Resource and Environmental Sciences, Wuhan University, Wuhan 430079, China (e-mail: \email{huifangli@whu.edu.cn}; \email{chaohu@whu.edu.cn}).}
\thanks{Shuang Luo is with Changjiang Spatial Information Technology Engineering Co., Ltd., Wuhan 430074, China (e-mail: \email{sluo@whu.edu.cn}).}
\thanks{Yan Luo is with the Department of Computing, The Hong Kong Polytechnic University, Hong Kong SAR, China (e-mail: \email{silver.luo@connect.polyu.hk}).}
\thanks{Chang Wen Chen is with the Department of Computing, The Hong Kong Polytechnic University, Hong Kong SAR, China, and also with Peng Cheng Laboratory, Shenzhen 518055, China (e-mail: \email{changwen.chen@polyu.edu.hk}).}
\thanks{Digital Object Identifier: 10.1109/TNNLS.2023.3336563}}

\markboth{IEEE TRANSACTIONS ON NEURAL NETWORKS AND LEARNING SYSTEMS}{Liu \textit{et al.}: Learning to Aggregate Multi-Scale Context for Instance Segmentation in Remote Sensing Images}

\maketitle

\begin{abstract}

The task of instance segmentation in remote sensing images, aiming at performing per-pixel labeling of objects at instance level, is of great importance for various civil applications. Despite previous successes, most existing instance segmentation methods designed for natural images encounter sharp performance degradations when they are directly applied to top-view remote sensing images. Through careful analysis, we observe that the challenges mainly come from the lack of discriminative object features due to severe scale variations, low contrasts, and clustered distributions. In order to address these problems, a novel context aggregation network (CATNet) is proposed to improve the feature extraction process. The proposed model exploits three lightweight plug-and-play modules, namely dense feature pyramid network (DenseFPN), spatial context pyramid (SCP), and hierarchical region of interest extractor (HRoIE), to aggregate global visual context at feature, spatial, and instance domains, respectively. DenseFPN is a multi-scale feature propagation module that establishes more flexible information flows by adopting inter-level residual connections, cross-level dense connections, and feature re-weighting strategy. Leveraging the attention mechanism, SCP further augments the features by aggregating global spatial context into local regions. For each instance, HRoIE adaptively generates RoI features for different downstream tasks. Extensive evaluations of the proposed scheme on iSAID, DIOR, NWPU VHR-10, and HRSID datasets demonstrate that the proposed approach outperforms state-of-the-arts under similar computational costs. Source code and pre-trained models are available at \href{https://github.com/yeliudev/CATNet}{https://github.com/yeliudev/CATNet}.

\end{abstract}


\section{Introduction}

\begin{figure}
\centering
\vspace{2mm}
\includegraphics[width=\linewidth]{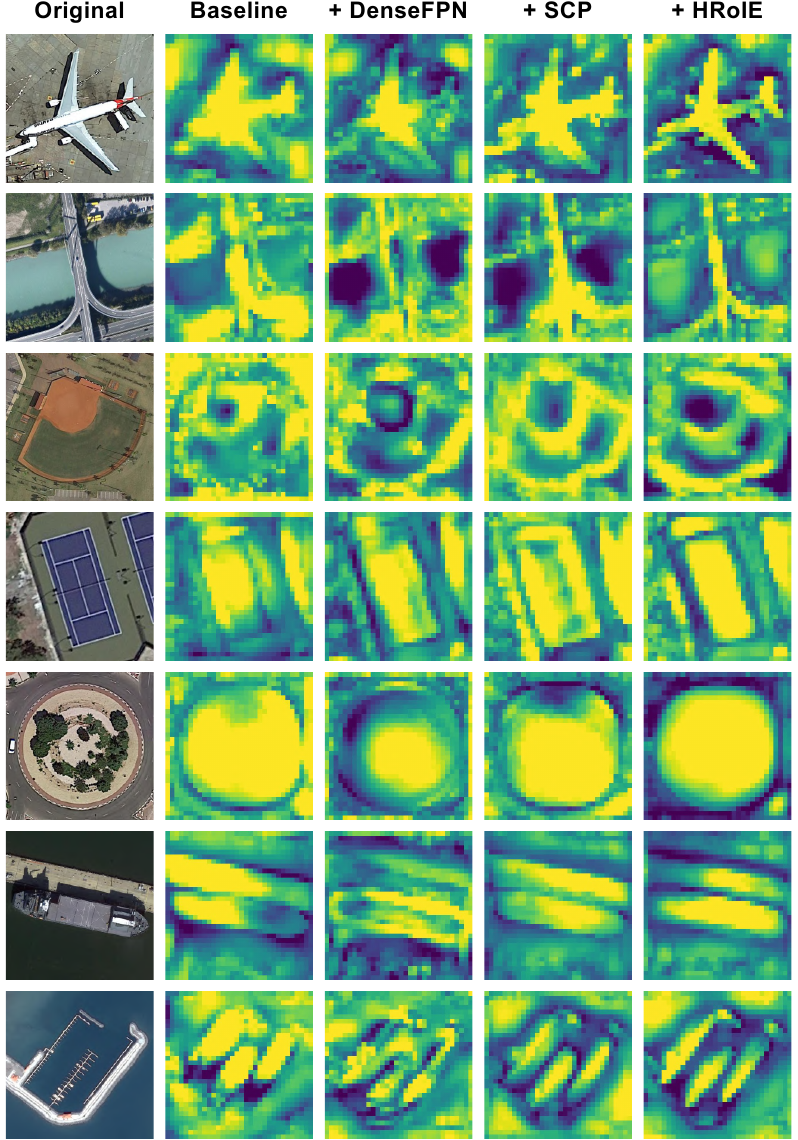}
\caption{Illustration of global context aggregation. Processed by the proposed modules, object features are enhanced by aggregating global visual context, as can be seen from the more discriminative feature maps.}
\label{fig:1}
\end{figure}

\IEEEPARstart{R}{ecent} advances in satellites and remote sensing techniques have generated a large variety of high-resolution remote sensing images, bringing great challenges to manual manipulation and processing. Therefore, automatic analysis and understanding of these images are becoming increasingly essential for various civil applications \cite{liu2022aerial,ding2022crowd,ding2022object,liu2022prompt,liu2023road,luo2023end}. As a fundamental yet challenging task in computer vision, instance segmentation, which is a combination of object detection and semantic segmentation aiming at predicting binary masks of objects at instance level, has been widely used to extract fine-grained object information from both optical remote sensing images and synthetic aperture radar (SAR) images. It has attracted considerable attention in recent years.

\begin{figure}
\centering
\vspace{2mm}
\includegraphics[width=\linewidth]{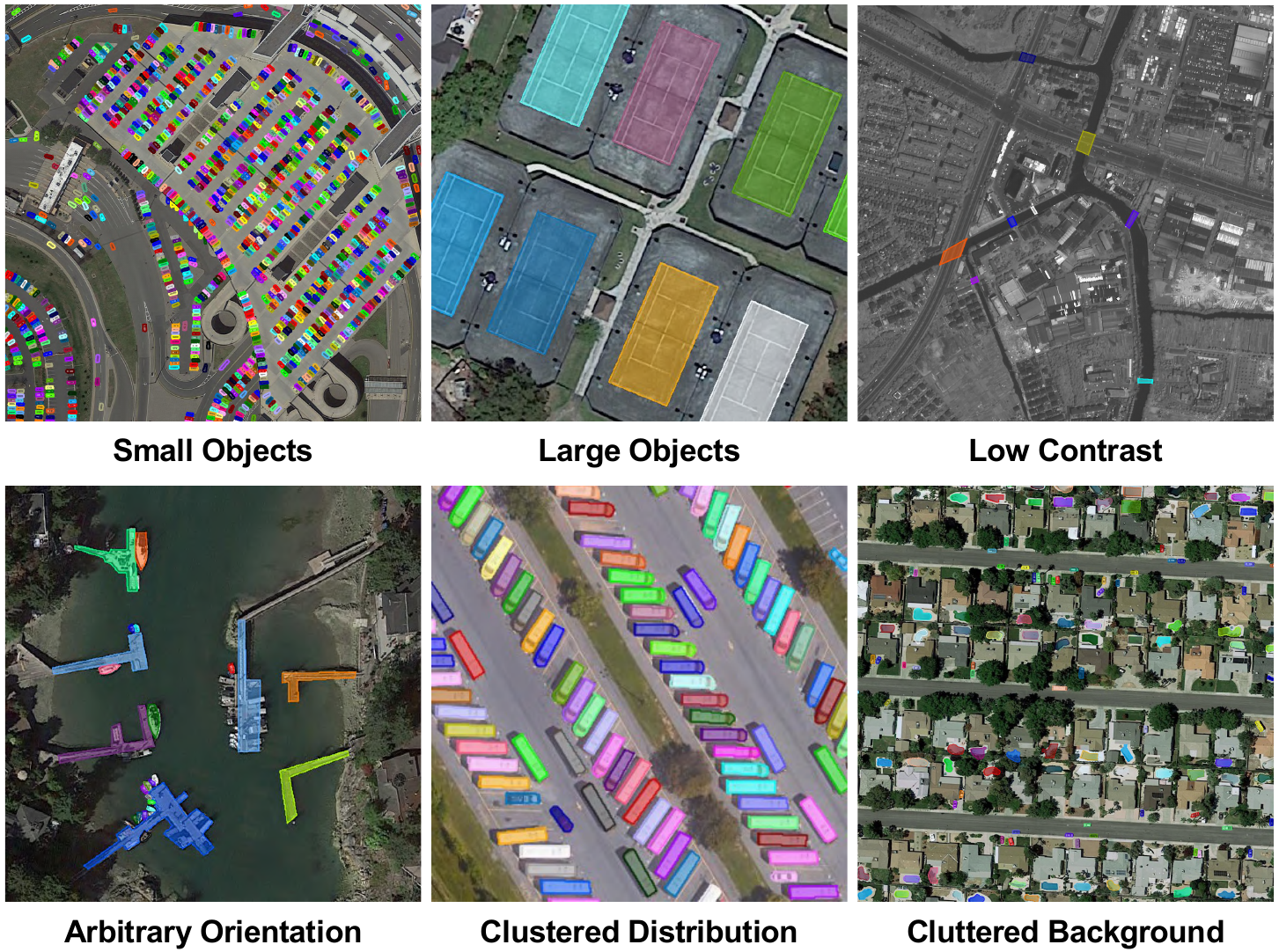}
\caption{The challenges of performing instance segmentation in remote sensing images. Here, scale variation, arbitrary orientation, and clustered distribution lead to complicated object patterns while low contrast and cluttered background bring interfering information from the background.}
\label{fig:2}
\end{figure}

Most existing works on object detection and instance segmentation \cite{ren2015faster,lin2017focal,he2017mask,chen2019hybrid,kirillov2019pointrend,wang2019solo,liu2022prompt} have been successful in conventional front-view scenes. However, when directly applied to remote sensing images, a large number of these methods inevitably encounter performance degradations \cite{cheng2014multi,xia2018dota,waqas2019isaid,li2020object,wei2020hrsid}. Compared with natural images, remote sensing images are typically viewed from the top, capturing large areas with limited object discrepancies. We analyze the peculiarities of these scenes and divide the challenges into five distinct aspects, \ie scale variation, arbitrary orientation, clustered distribution, low contrast, and cluttered background, as illustrated in Fig.~\ref{fig:2}. The first three aspects lead to complicated object patterns while the remaining two bring interfering information from the background. These phenomena are scarce in natural scenes, thus only a few works have considered these aspects. We argue that all the challenges above are due to the lack of discriminative object features in remote sensing images. That is, visual appearances of individual objects in remote sensing images are not informative enough for directly adopting existing schemes to perform instance segmentation. Such a view is also supported by previous works \cite{yang2020scrdet++,zhang2020contextual,cheng2020cross,lin2020novel,zeng2021cpisnet}.

A natural question will be: \textit{How to enhance the inadequate features to achieve better instance segmentation results in remote sensing images?} Considering that, in a general instance segmentation pipeline \cite{he2017mask,chen2019hybrid,kirillov2019pointrend}, object representations are directly cropped from either of the feature maps from the backbone or neck, containing only local features with irreversible information loss. In this work, we mitigate this problem by introducing CATNet, a novel framework for global context aggregation. The key idea is that context information of images, coming from different feature pyramid levels, spatial positions, or receptive fields, shall provide extra prior for segmenting indistinguishable objects. Note that existing works \cite{wang2018non,zhang2019cad} only regard context as spatial correlations. We expand and explicitly disentangle the concept of context into three domains, \ie \textit{feature}, \textit{space}, and \textit{instance}. The implication is that when detecting and segmenting objects, this model may augment the visual information by: 1) balancing the heterogeneous features, 2) fusing information from the background or other correlative objects, and 3) adaptively refining intermediate representations for each instance and task. These three different domains are capable of modeling global visual context from coarse to fine at different granularities, capturing more discriminative object information.

The proposed framework intends to leverage three plug-and-play modules and construct the aforementioned context aggregation pipeline. Fig.~\ref{fig:1} shows a glimpse of object features processed by these modules. In the feature domain, we argue that, in the feature pyramid built up by a backbone, flexible information flows may reduce information confounding and handle multi-scale features more effectively. It is based on this analysis that a dense feature pyramid network (DenseFPN) is proposed to enable adaptive feature propagation. This module has a pyramid structure with stackable basic blocks consisting of top-down and bottom-up paths. We adopt inter-level residual connections \cite{he2016deep}, cross-level dense connections \cite{huang2017densely}, and feature re-weighting strategy to enable the module to learn its optimal feature propagation manner. In the spatial domain, long-range spatial dependencies in remote sensing images bring more complementary information to blurry objects than in natural scenes. So the spatial context pyramid (SCP) is adopted to capture global spatial context in each feature pyramid level. This module learns to aggregate features from the whole feature map, and combines them into each pixel using adaptive weights. Such a strategy guarantees that only useful global information is fused into local regions, without decreasing the discrepancies among objects. As for the instance domain, we argue that object representations should be adaptively refined for each instance and downstream task. For example, performing object classification needs an overall view, while segmentation requires more zoomed details. The demand for different sizes of receptive fields also varies among instances. Hence, we introduce hierarchical region of interest extractor (HRoIE) to generate RoI features per instance and task. After cropping the instance feature maps from all levels, this module starts from the highest or lowest scale, and fuses the features level-by-level in a hierarchical manner. Pixel-wise attention mechanism is exploited to combine neighboring feature maps. These modules are lightweight while having the flexibility for scalable model design. Overall, the main contributions of this paper are three-fold:

\begin{figure*}
\centering
\includegraphics[width=\linewidth]{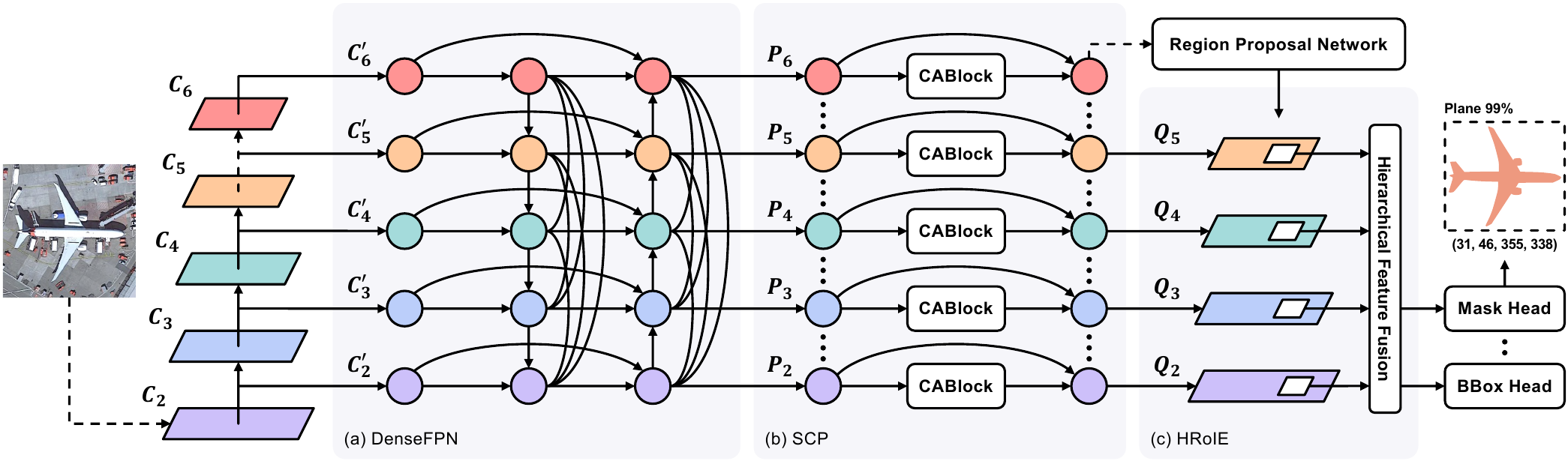}
\caption{Overall architecture of the proposed framework. The process of global context aggregation is realized by three modules, namely a) dense feature pyramid network, b) spatial context pyramid, and c) hierarchical region of interest extractor. These modules are designed to aggregate global context information from different feature pyramid levels, spatial positions, and receptive fields at feature, spatial, and instance domains, respectively.}
\label{fig:3}
\end{figure*}

\begin{itemize}
\item The expansion and explicit disentangling of the concept of context into \textit{feature}, \textit{spatial}, and \textit{instance} domains have resulted in superior performance in both optical remote sensing image and SAR image segmentation. To the best of our knowledge, this is the first work that considers global visual context beyond spatial dependencies.
\item The proposed CATNet is capable of utilizing DenseFPN, SCP, and HRoIE to learn and aggregate the global visual context from different domains for object detection and instance segmentation in remote sensing images.
\item The proposed scheme has been tested on a wide variety of datasets, including iSAID, DIOR, NWPU VHR-10, and HRSID, and new state-of-the-art performance has been obtained with similar computational costs.
\end{itemize}

The rest of this paper is organized as follows. Related works and comparisons are discussed in Section~\ref{sec:2}. Detailed formulations of DenseFPN, SCP, and HRoIE are introduced in Section~\ref{sec:3}. Section~\ref{sec:4} presents extensive experimental results and in-depth analysis on public datasets. Finally, concluding remarks are summarized in Section~\ref{sec:5}.

\section{Related Work}\label{sec:2}

\subsection{Instance Segmentation in Remote Sensing Images}

Instance Segmentation is a challenging and broadly studied problem in computer vision. Similar to object detection \cite{ren2015faster,lin2017focal}, the majority of instance segmentation approaches can be divided into two schemes, namely one-stage methods and two-stage methods. As a straightforward design, one-stage methods \cite{bolya2019yolact,wang2019solo} adopt the bottom-up strategy that performs semantic segmentation at image level, and further separates individual objects using clustering or metric learning. These methods often possess considerable efficiencies, but are largely restricted by their localization accuracy. Compared to this paradigm, two-stage methods \cite{he2017mask,liu2018path,chen2019hybrid,kirillov2019pointrend} separate the segmentation pipeline into two phases, \ie region proposal generation and task-specific post-processing, resulting in a top-down style. Benefiting from two-time bounding box regression, these methods usually achieve better results on object localization and mask prediction. Some recent works \cite{he2019skip,wang2020looking,zhao2021mgml,ma2021multi,lin2020crpn,zhang2020contextual,cheng2020cross,lin2020novel,wang2021fsod,yang2020scrdet++} try to tackle the problem of scene classification and object detection in remote sensing images, but they do not pay particular attention to instance segmentation. Our proposed context aggregation strategy can be integrated into both one-stage and two-stage methods, while HRoIE is not used in one-stage methods since it's not necessary to crop the feature maps. Further experimental results demonstrate that our modules can steadily boost performances.

\begin{figure}
\centering
\includegraphics[width=0.95\linewidth]{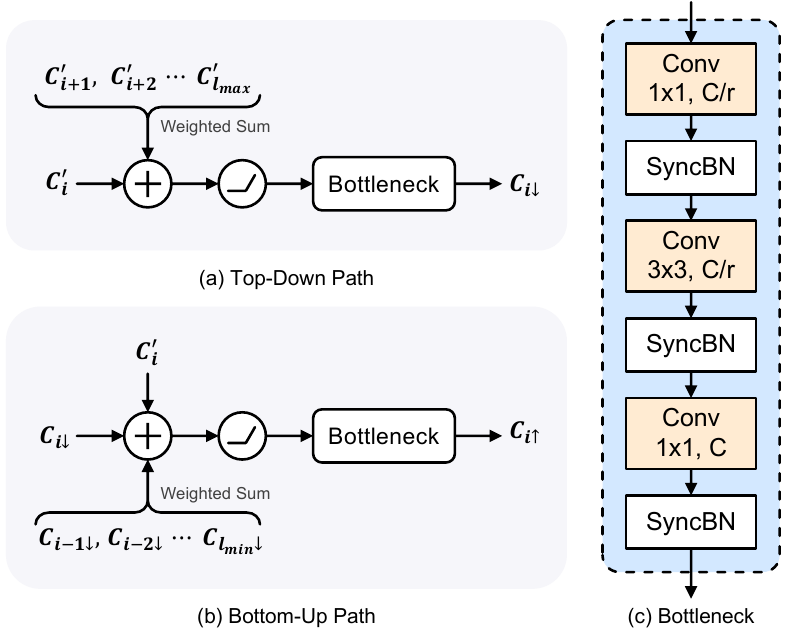}
\caption{Detailed procedure of multi-scale feature propagation in dense feature pyramid network. $\oplus$ and $\orelu$ denote element-wise addition and ReLU activation, respectively. The cross-level feature maps are adaptively combined using a feature re-weighting strategy.}
\label{fig:4}
\end{figure}

\begin{figure*}
\centering
\includegraphics[width=\linewidth]{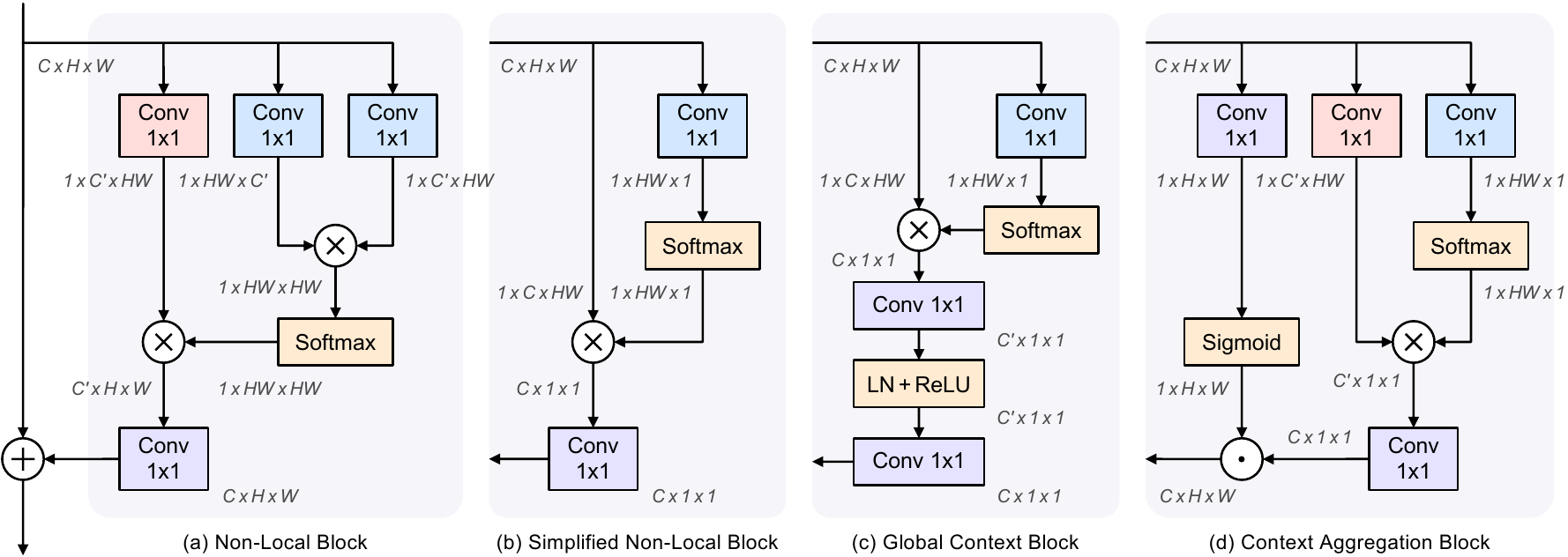}
\caption{Designs of spatial context modules. Permutations of feature maps are represented as their dimensions, \eg $C \! \times \! H \! \times \! W$ indicates a matrix with number of channels $C$, height $H$, and width $W$. $\otimes$, $\odot$, and $\oplus$ denote batched matrix multiplication, broadcast hadamard product, and broadcast element-wise addition, respectively. Convolution layers used for attention map generation, feature mapping, and context refinement are annotated as blue, red, and purple.}
\label{fig:5}
\end{figure*}

\subsection{Multi-Scale Feature Propagation}

Scale variation is a long-standing challenge in most visual recognition tasks, and it is a common solution to leverage a pyramid structure to represent the visual features under different resolutions. In the area of dense prediction, FPN \cite{lin2017feature} is the first work that builds up a feature pyramid, and propagates the information among different levels. Such a design rapidly became a standard for most instance segmentation models. However, FPN only propagates the features using a top-down path, which is suboptimal for multi-level feature fusion. Thus PAFPN \cite{liu2018path} was proposed to incorporate an extra bottom-up propagation path, and FPG \cite{chen2020feature} further introduced a multi-pathway feature pyramid that can better capture cross-level information. MHN \cite{cao2019high} tackles the semantic gap problem by leveraging semantic feature maps. A similar multi-branch structure is also used in TridentNet \cite{li2019scale} and NETNet \cite{li2020netnet} for scale-aware training and generating scale-aware features. From the perspective of basic operators, scale-equalizing pyramid convolution (SEPC) \cite{wang2020scale} tends to fuse the feature maps using 3D convolutions. More recently, recursive feature pyramid (RFP) \cite{qiao2021detectors} re-uses the backbone to capture deeper semantics via feedback connections from FPN. In order to reduce the computational costs, NAS-FPN \cite{ghiasi2019nas} is obtained by introducing neural architecture search (NAS) \cite{zoph2018learning} to optimize the architecture of FPN, and BiFPN \cite{tan2019efficientdet} is constructed by carefully designing the feature fusion blocks. We argue that compared with hard-coded feature aggregation paths, flexible information flows may reduce confounding and handle multi-scale features more effectively, so that DenseFPN is proposed to learn the optimal feature aggregation strategy during training.

\subsection{Global Context Modeling}

One of the most representative properties of convolutional neural networks (CNNs) is local dependency modeling. The receptive fields can only be enlarged by stacking multiple convolution layers. Although feature pyramids have been introduced to capture multi-scale information, long-range spatial dependency modeling has also been proven to be effective for dense prediction tasks \cite{hu2018squeeze,wang2018non}. As a pathbreaking work, non-local neural network (NLNet) \cite{wang2018non} shows that global spatial context can be aggregated by computing pixel-level pairwise correlations, but it suffers from the problem of high computational cost. Therefore, some extensions of NLNet tackle this problem by simplifying the pairwise correlation computation. For example, criss-cross attention (CCNet) \cite{huang2019ccnet} aggregates global context by fusing the information along axes twice. Xu \etal observed that the per-pixel attention maps in NLNet are almost the same for different positions, so that GCNet \cite{cao2023global} is proposed to generate a single attention map for a feature map. GANet \cite{zhu2019empirical} acts as a unified attention module for global context modeling. Although promising results have been achieved by these works, all these methods merely consider context as long-range spatial correlations, ignoring the dependencies in feature and instance domains.

\section{The Proposed Approach}\label{sec:3}

In this section, we introduce our approach on global context aggregation. As shown in Fig.~\ref{fig:3}, the entire framework can be divided into three sub-modules, namely DenseFPN, SCP, and HRoIE. These modules aim to aggregate global context information in order from feature, spatial, and instance domains, respectively.

\subsection{Overview}

Given an image $x$ and a set of object categories of interest $S = \{1,...,N\}$, the task of instance segmentation aims to detect and segment all the objects in $x$, where they belong to whichever the pre-defined categories. The output of instance segmentation would be a collection of tuples $\mathcal{T} = \{\langle b, m, s \rangle\}$, where $b \in \mathbb{R}^4$ denotes the bounding box of the object, $m$ represents a binary mask in which $m_{i, j} \in \{0, 1\}$ indicates whether the pixel $(i, j)$ belongs to this object, and $s \in S$ is a one-hot vector describing the object category. Note that a single object may be presented by separate masks.

We adopt Mask R-CNN \cite{he2017mask}, a common two-stage instance segmentation framework, as our baseline. The whole pipeline is constructed by extracting visual features, generating region proposals, and performing bounding box regression, object classification as well as mask prediction on each proposal. A heterogeneous feature pyramid is first built by extracting visual features from each stage of the backbone. In order to make the features more discriminative, we exploit DenseFPN and SCP to propagate object information among different levels and regions. After enhancing the feature pyramid, task-specific RoI features are generated by HRoIE for each proposal. Details of these modules are introduced in the following sections.

\subsection{Dense Feature Pyramid Network}

Multi-scale feature propagation aims to aggregate visual features from different backbone stages, that is given an input feature pyramid $C = \{C_{l_1}, C_{l_2}, ...\}$, where $C_i$ denotes the feature map from the stage $i$, the goal is to propagate the features among different levels to produce an enhanced feature pyramid $P = \{P_{l_1}, P_{l_2}, ...\}$, in which the features are more informative for downstream tasks. Formally, the resolution of feature map $C_i$ or $P_i$ is $1 / 2^i$ of the input image.

\begin{figure}
\centering
\vspace{2.5mm}
\includegraphics[height=6.5cm]{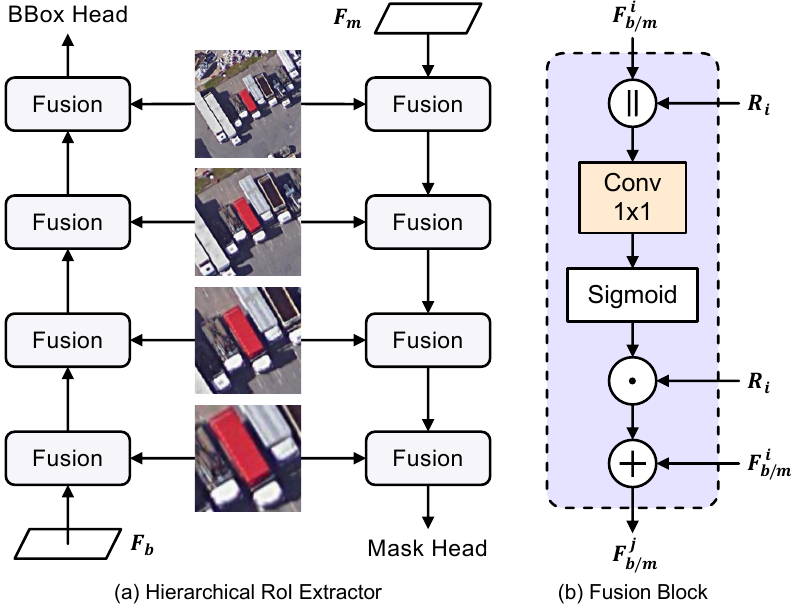}
\caption{Detailed structure of hierarchical region of interest extractor. $\oparallel$, $\odot$, and $\oplus$ denote channel-wise concatenation, hadamard product, and element-wise addition, respectively. The RoI features are generated by fusing multi-scale feature maps along top-down or bottom-up paths in a hierarchical manner.}
\label{fig:6}
\end{figure}

The basic architecture of DenseFPN is shown in Fig.~\ref{fig:3}~(a), where each node represents a feature map and the lines stand for information flows. This module takes $C_2 \sim C_5$ as inputs and firstly down-samples them to $256$ channels using $1 \times 1$ convolutions, producing $C_2' \sim C_5'$. An extra $3 \times 3$ convolution with $stride = 2$ is applied to $C_5'$ to generate $C_6'$. So that $C_2' \sim C_6'$ are with the same number of channels but different resolutions. Subsequently, these features are passed through several stacked basic blocks for feature-level context aggregation. In each block, the input feature pyramid is processed by a top-down and a bottom-up aggregation path, in which inter-level residual connections \cite{he2016deep}, cross-level dense connections \cite{huang2017densely}, and feature re-weighting strategy are adopted.

\begin{table*}[h!]
\caption{Object detection and instance segmentation results on iSAID test set.}
\label{tab:1}
\renewcommand\tabcolsep{0pt}
\footnotesize
\begin{threeparttable}
\begin{tabularx}{\linewidth}{@{\hspace{0.15cm}}p{2.85cm}<{\raggedright}|@{\hspace{0.15cm}}p{1.75cm}<{\raggedright}|p{1.1cm}<{\centering}p{1.1cm}<{\centering}p{1.1cm}<{\centering}p{1.1cm}<{\centering}p{1.1cm}<{\centering}p{1.1cm}<{\centering}|p{1.1cm}<{\centering}p{1.1cm}<{\centering}p{1.1cm}<{\centering}p{1.1cm}<{\centering}p{1.1cm}<{\centering}p{1.1cm}<{\centering}}
\toprule
\textbf{Method}&\textbf{Backbone}&\textbf{AP}\boldmath{$_b$}&\textbf{AP}\boldmath{$_b^{50}$}&\textbf{AP}\boldmath{$_b^{75}$}&\textbf{AP}\boldmath{$_b^S$}&\textbf{AP}\boldmath{$_b^M$}&\textbf{AP}\boldmath{$_b^L$}&\textbf{AP}\boldmath{$_m$}&\textbf{AP}\boldmath{$_m^{50}$}&\textbf{AP}\boldmath{$_m^{75}$}&\textbf{AP}\boldmath{$_m^S$}&\textbf{AP}\boldmath{$_m^M$}&\textbf{AP}\boldmath{$_m^L$}\\
\midrule
\multirow{2}{*}{Mask R-CNN \cite{he2017mask}} & ResNet-50 & 42.2 & 63.1 & 47.8 & 44.7 & 49.9 & 18.9 & 36.7 & 59.7 & 39.7 & 39.5 & 45.1 & 11.5 \\
& ResNet-101 & 43.9 & 65.5 & 49.7 & 46.4 & 53.4 & 18.8 & 38.3 & 61.7 & 40.9 & 41.0 & 48.4 & 13.4 \\
\midrule
\multirow{2}{*}{CenterMask \cite{lee2020centermask}} & ResNet-50 & 44.6 & 65.6 & 49.7 & 47.4 & 48.7 & 21.4 & 37.4 & 61.4 & 39.4 & 40.2 & 43.6 & 15.0 \\
& ResNet-101 & 44.4 & 65.6 & 49.2 & 47.5 & 48.2 & 17.3 & 37.2 & 61.0 & 38.9 & 40.3 & 42.7 & 11.9 \\
\midrule
\multirow{2}{*}{BlendMask \cite{chen2020blendmask}} & ResNet-50 & 45.1 & 66.1 & 50.1 & 47.5 & 50.6 & 21.9 & 38.2 & 62.4 & 40.0 & 40.6 & 46.9 & 18.0 \\
& ResNet-101 & 45.2 & 65.7 & 50.1 & 47.9 & 51.6 & 22.9 & 38.3 & 62.5 & 40.0 & \textbf{48.9} & 47.4 & 17.9 \\
\midrule
\multirow{2}{*}{DB-BlendMask \cite{chen2021db}} & ResNet-50 & 46.0 & 66.8 & 50.8 & 48.8 & 50.7 & 21.2 & 39.2 & 63.5 & 41.4 & 42.0 & 46.5 & 16.3 \\
& ResNet-101 & 46.0 & 66.8 & 50.8 & 48.8 & 52.3 & 22.8 & 38.9 & 63.1 & 41.0 & 41.6 & 48.1 & 17.8 \\
\midrule
Cascade R-CNN \cite{cai2018cascade} & ResNeXt-101 & 45.3 & 65.8 & 51.1 & 48.3 & 53.7 & 22.8 & 37.0 & 60.7 & 39.0 & 39.8 & 46.6 & 15.1 \\
HTC \cite{chen2019hybrid} & ResNet-152 & 46.3 & 66.4 & 51.8 & 48.5 & 56.4 & 25.3 & 38.3 & 61.6 & 40.9 & 40.6 & 50.0 & 17.4 \\
ISDNet \cite{garg2021isdnet} & ResNet-152 & 46.8 & 67.1 & 52.4 & 49.0 & 57.4 & 27.3 & 38.7 & 62.2 & 42.2 & 41.1 & 51.0 & 19.8 \\
MS R-CNN \cite{huang2019mask} & ResNet-101 & -- & -- & -- & -- & -- & -- & 37.0 & 57.8 & 40.5 & 39.7 & 46.0 & 14.3 \\
SCNet \cite{vu2021scnet} & ResNet-101 & -- & -- & -- & -- & -- & -- & 38.1 & 60.4 & 41.2 & 40.9 & 46.9 & 12.6 \\
CPISNet \cite{zeng2021cpisnet} & AFEN-4GF & -- & -- & -- & -- & -- & -- & 39.1 & 62.2 & 42.5 & 41.8 & 49.6 & 17.6 \\
\midrule
\textbf{CATNet} & ResNet-50 & 48.0 & 66.1 & 53.9 & 51.3 & 57.6 & 22.6 & 39.9 & 62.8 & 43.5 & 43.1 & 50.2 & 13.4 \\
\textbf{CATNet} + Aug. & ResNet-50 & \textbf{50.7} & \textbf{69.4} & \textbf{57.1} & \textbf{54.3} & \textbf{59.9} & \textbf{33.8} & \textbf{41.9} & \textbf{66.0} & \textbf{45.5} & 45.2 & \textbf{52.8} & \textbf{25.7} \\
\bottomrule
\end{tabularx}
\vspace{1.3mm}
\hspace{0.5mm}
AP$_b$\,-\,Bounding box\,AP,\, AP$_m$\,-\,Mask\,AP,\, Aug.\,-\,Multi-Scale\,Training
\end{threeparttable}
\end{table*}

\begin{table*}
\caption{Class-wise instance segmentation results on iSAID test set.}
\label{tab:2}
\renewcommand\tabcolsep{0pt}
\footnotesize
\begin{threeparttable}
\begin{tabularx}{\linewidth}{@{\hspace{0.15cm}}p{2.65cm}<{\raggedright}|@{\hspace{0.15cm}}p{1.675cm}<{\raggedright}|p{0.95cm}<{\centering}@{\hspace{-0.15cm}}p{0.95cm}<{\centering}|@{\hspace{0.075cm}}p{0.775cm}<{\centering}p{0.775cm}<{\centering}p{0.775cm}<{\centering}p{0.775cm}<{\centering}p{0.775cm}<{\centering}p{0.775cm}<{\centering}p{0.775cm}<{\centering}p{0.775cm}<{\centering}p{0.775cm}<{\centering}p{0.775cm}<{\centering}p{0.775cm}<{\centering}p{0.775cm}<{\centering}p{0.775cm}<{\centering}p{0.775cm}<{\centering}p{0.775cm}<{\centering}}
\toprule
\textbf{Method}&\textbf{Backbone}&\textbf{AP}\boldmath{$_b$}&\textbf{AP}\boldmath{$_m$}&\textbf{SH}&\textbf{ST}&\textbf{BD}&\textbf{TC}&\textbf{BC}&\textbf{GT}&\textbf{BR}&\textbf{LV}&\textbf{SV}&\textbf{HE}&\textbf{SP}&\textbf{RO}&\textbf{SB}&\textbf{PL}&\textbf{HA}\\
\midrule
\multirow{2}{*}{Mask R-CNN \cite{he2017mask}} & ResNet-50 & 42.2 & 36.7 & 46.8 & 35.4 & 48.0 & 75.7 & 48.2 & 28.1 & 17.4 & 30.4 & 15.6 & 13.4 & 38.1 & 43.2 & 30.9 & 42.9 & 31.0 \\
& ResNet-101 & 43.9 & 38.3 & 47.0 & 32.0 & 53.1 & 76.7 & 48.2 & 30.4 & 18.6 & 31.1 & 16.1 & 14.5 & 39.0 & 46.4 & 35.7 & 44.1 & 31.1 \\
\midrule
\multirow{2}{*}{CenterMask \cite{lee2020centermask}} & ResNet-50 & 44.6 & 37.4 & 46.9 & 37.7 & 54.0 & 76.7 & 50.5 & 23.6 & 16.7 & 32.5 & 16.1 & 10.0 & 38.1 & 41.5 & 35.9 & 42.2 & 29.9 \\
& ResNet-101 & 44.4 & 37.2 & 46.4 & 36.9 & 54.2 & 75.8 & 50.1 & 25.7 & 16.5 & 32.2 & 15.6 & 10.5 & 37.2 & 42.2 & 34.6 & 41.2 & 29.9 \\
\midrule
\multirow{2}{*}{BlendMask \cite{chen2020blendmask}} & ResNet-50 & 45.1 & 38.2 & 46.9 & 37.4 & 55.5 & 76.4 & 50.9 & 21.3 & 15.9 & 33.2 & 15.5 & 11.0 & 38.4 & 45.3 & 38.6 & 43.8 & 33.5 \\
& ResNet-101 & 45.2 & 38.3 & 47.1 & 38.0 & 56.4 & 77.3 & 51.2 & 22.4 & 16.7 & 33.4 & 16.2 & 11.8 & 38.2 & 45.5 & 36.0 & 42.6 & 34.0 \\
\midrule
\multirow{2}{*}{DB-BlendMask \cite{chen2021db}} & ResNet-50 & 46.0 & 39.2 & 48.6 & 38.2 & 56.3 & 78.3 & 52.2 & 23.7 & 17.6 & 33.6 & \textbf{16.4} & 12.4 & 38.1 & 44.7 & 36.9 & 44.7 & \textbf{37.4} \\
& ResNet-101 & 46.0 & 38.9 & 48.6 & 37.3 & 54.8 & 77.5 & 51.2 & 23.8 & 17.8 & \textbf{34.1} & 16.3 & 12.2 & 38.4 & 44.4 & 36.3 & 45.2 & 36.1 \\
\midrule
Cascade R-CNN \cite{cai2018cascade} & ResNeXt-101 & 45.3 & 37.0 & 46.7 & 36.7 & 54.7 & 76.4 & 52.7 & 20.6 & 18.1 & 30.5 & 14.1 & 11.4 & 39.3 & 45.8 & 37.3 & 41.8 & 28.4 \\
HTC \cite{chen2019hybrid} & ResNet-152 & 46.3 & 38.3 & 47.3 & 37.9 & 53.5 & 76.4 & 51.0 & 32.5 & 19.0 & 31.2 & 16.3 & 11.4 & 38.4 & 45.6 & 39.6 & 43.1 & 30.7 \\
ISDNet \cite{garg2021isdnet} & ResNet-152 & 46.8 & 38.7 & 47.8 & 38.7 & 54.0 & 76.3 & 52.4 & 32.8 & 19.3 & 31.8 & 16.6 & 12.0 & 39.0 & 45.0 & 40.1 & 43.3 & 31.0 \\
MS R-CNN \cite{huang2019mask} & ResNet-101 & -- & 37.0 & 46.6 & 33.9 & 54.2 & 76.1 & 49.9 & 29.7 & 17.7 & 30.0 & 14.0 & 11.8 & 37.9 & 44.4 & 35.7 & 43.8 & 30.1 \\
SCNet \cite{vu2021scnet} & ResNet-101 & -- & 38.1 & 48.0 & 35.9 & 56.6 & 77.0 & 51.5 & 30.2 & 18.7 & 31.7 & 14.2 & 9.7 & 39.2 & 46.7 & 36.3 & 45.1 & 31.1 \\
CPISNet \cite{zeng2021cpisnet} & AFEN-4GF & -- & 39.1 & \textbf{49.5} & 35.8 & 54.3 & 77.6 & 52.9 & 31.9 & 19.9 & 32.9 & 14.9 & 13.2 & \textbf{40.6} & 45.2 & 39.3 & 46.2 & 32.7 \\
\midrule
\textbf{CATNet} & ResNet-50 & 48.0 & 39.9 & 49.4 & 36.7 & 55.9 & 77.7 & 55.9 & 29.7 & 20.0 & 31.5 & 15.1 & 14.8 & 40.1 & 43.5 & 39.0 & \textbf{47.0} & 35.1 \\
\textbf{CATNet} + Aug. & ResNet-50 & \textbf{50.7} & \textbf{41.9} & 48.9 & \textbf{39.4} & \textbf{60.9} & \textbf{79.2} & \textbf{56.7} & \textbf{33.8} & \textbf{21.2} & 31.6 & 15.6 & \textbf{17.8} & 39.7 & \textbf{48.6} & \textbf{45.8} & 46.6 & 36.5 \\
\bottomrule
\end{tabularx}
\vspace{1.3mm}
\hspace{0.5mm}
AP$_b$\,-\,Bounding box\,AP,\, AP$_m$\,-\,Mask\,AP,\, Aug.\,-\,Multi-Scale\,Training
\end{threeparttable}
\end{table*}

Fig.~\ref{fig:4} illustrates the detailed feature propagation strategy in basic blocks. In the top-down path, output features $C_{i\downarrow}$ of each feature pyramid level are generated by fusing the features from the current level and all upper levels, then performing a parameterized transform upon the fused features.
\begin{gather}\label{eq:1}
C_{i\downarrow} = Bottleneck(C_i' + \hspace{-1.5mm} \sum_{j=i+1}^{l_{max}} [Resize(C_j') \cdot w_{i\downarrow}^j])
\end{gather}
Here, $Bottleneck(\cdot)$ denotes a ReLU activation layer followed by a $3 \times 3$ bottleneck \cite{he2016deep} without activations. We observe that adopting only one nonlinearity before the bottleneck structure brings better performance. $Resize(\cdot)$ represents a max pooling layer, and $w_{i\downarrow}^j$ is a learnable re-weighting term for aggregating features from level $j$ to level $i$. The weights $w_{i\downarrow}$ are vectors with lengths correspond to their levels, the values are normalized from raw values using softmax by
\begin{gather}
w_{i\downarrow}^j = \frac{\exp(v_{i\downarrow}^j)}{\sum_{k=1}^{N_i} \exp(v_{i\downarrow}^k)}
\end{gather}
where $v_{i\downarrow}$ denotes the raw weight vector and $j$ is the index of each element. Using the normalization above can stabilize the learning process. Similarly to the top-down path, bottom-up features $C_{2\uparrow} \sim C_{6\uparrow}$ are computed by
\begin{gather}
C_{i\uparrow} = Bottleneck(C_i' + C_{i\downarrow} + \hspace{-2mm} \sum_{j=l_{min}}^{i-1} \hspace{-0.5mm} [Resize(C_{j\downarrow}) \cdot w_{i\uparrow}^j])
\end{gather}
where $Resize(\cdot)$ represents a bilinear interpolation layer and other notations are consistent with Eq.~\ref{eq:1}. We adopt residual connections to preserve the original features and prevent gradient vanishing. Leveraging the flexible architecture and feature re-weighting strategy, DenseFPN has the capability to optimize the information flow of context aggregation in feature domain during training.

\begin{figure}
\centering
\vspace{2.5mm}
\includegraphics[height=6.5cm]{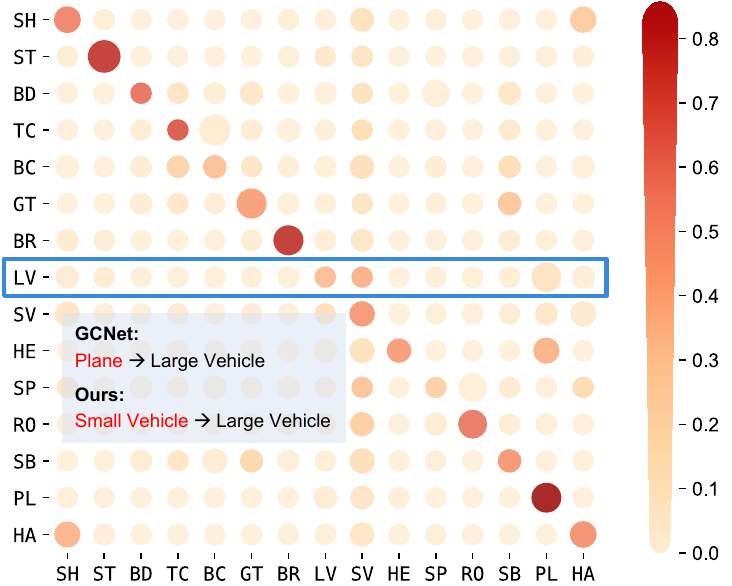}
\caption{Visualization of feature aggregation weights in iSAID dataset. X-axis refers to the categories that used to construct the global context, while the y-axis denotes the categories that receive features. The weights in GCNet and SCP are represented as sizes and depths of colors, respectively.}
\label{fig:7}
\end{figure}

\subsection{Spatial Context Pyramid}

After aggregating feature maps across different levels, the feature pyramid remains to contain spatially local information, thus we introduce spatial context pyramid (SCP) to further augment the features by learning global spatial context within each level. Former attempts in this area \cite{hu2018squeeze,wang2018non,zhu2019empirical,huang2019ccnet,cao2023global} normally integrate several spatial or channel attention blocks into the backbone to enable global receptive fields. Some architectures of these blocks are presented in Fig.~\ref{fig:5}. Among these methods, non-local neural network (NLNet) \cite{wang2018non} is a general solution that computes pixel-wise correlations via embedded gaussian for every spatial position, while squeeze-and-excitation network (SENet) \cite{hu2018squeeze} tackles the problem from the perspective of channel attentions. To combine the spatial and channel modulation abilities, global context network (GCNet) \cite{cao2023global} is proposed to learn a single attention map for an image. However, we observe that in remote sensing images with objects only covering small areas, this design may bring too much useless background information to objects. To tackle this problem, we propose adding an extra path on top of this structure to learn the informativeness of each pixel. Our core idea is that if the features of a pixel are informative enough, there's not much need to aggregate features from other spatial positions. Such a soft re-weighting strategy can effectively fuse global features while reducing information confounding.

\begin{table*}
\renewcommand\tabcolsep{0pt}
\caption{Class-wise object detection results on DIOR test set.}
\label{tab:3}
\footnotesize
\begin{threeparttable}
\begin{tabularx}{\linewidth}{@{\hspace{0.15cm}}p{2.4cm}<{\raggedright}|@{\hspace{0.15cm}}p{1.5cm}<{\raggedright}|p{0.75cm}<{\centering}|@{\hspace{0.05cm}}p{0.65cm}<{\centering}p{0.65cm}<{\centering}p{0.65cm}<{\centering}p{0.65cm}<{\centering}p{0.65cm}<{\centering}p{0.65cm}<{\centering}p{0.65cm}<{\centering}p{0.65cm}<{\centering}p{0.65cm}<{\centering}p{0.65cm}<{\centering}p{0.65cm}<{\centering}p{0.65cm}<{\centering}p{0.65cm}<{\centering}p{0.65cm}<{\centering}p{0.65cm}<{\centering}p{0.65cm}<{\centering}p{0.65cm}<{\centering}p{0.65cm}<{\centering}p{0.65cm}<{\centering}p{0.65cm}<{\centering}}
\toprule
\textbf{Method}&\textbf{Backbone}&\textbf{AP}\boldmath{$_b$}&\textbf{AL}&\textbf{AR}&\textbf{BF}&\textbf{BC}&\textbf{BR}&\textbf{CH}&\textbf{DA}&\textbf{ES}&\textbf{ET}&\textbf{GC}&\textbf{GT}&\textbf{HA}&\textbf{OV}&\textbf{SH}&\textbf{ST}&\textbf{SA}&\textbf{TC}&\textbf{TS}&\textbf{VE}&\textbf{WM}\\
\midrule
R-CNN \cite{girshick2014rich} & VGG-16 & 37.7 & 35.6 & 43.0 & 53.8 & 62.3 & 15.6 & 53.7 & 33.7 & 50.2 & 33.5 & 50.1 & 49.3 & 39.5 & 30.9 & 9.1 & 60.8 & 18.0 & 54.0 & 36.1 & 9.1 & 16.4 \\
RICNN \cite{cheng2016learning} & VGG-16 & 44.2 & 39.1 & 61.0 & 60.1 & 66.3 & 25.3 & 63.3 & 41.1 & 51.7 & 36.6 & 55.9 & 58.9 & 43.5 & 39.0 & 9.1 & 61.1 & 19.1 & 63.5 & 46.1 & 11.4 & 31.5 \\
RICAOD \cite{li2017rotation} & VGG-16 & 50.9 & 42.2 & 69.7 & 62.0 & 79.0 & 27.7 & 68.9 & 50.1 & 60.5 & 49.3 & 64.4 & 65.3 & 42.3 & 46.8 & 11.7 & 53.5 & 24.5 & 70.3 & 53.3 & 20.4 & 56.2 \\
RIFD-CNN \cite{cheng2016rifd} & VGG-16 & 56.1 & 56.6 & 53.2 & 79.9 & 69.0 & 29.0 & 71.5 & 63.1 & 69.0 & 56.0 & 68.9 & 62.4 & 51.2 & 51.1 & 31.7 & 73.6 & 41.5 & 79.5 & 40.1 & 28.5 & 46.9 \\
SSD \cite{liu2016ssd} & VGG-16 & 58.6 & 59.5 & 72.7 & 72.4 & 75.7 & 29.7 & 65.8 & 56.6 & 63.5 & 53.1 & 65.3 & 68.6 & 49.4 & 48.1 & 59.2 & 61.0 & 46.6 & 76.3 & 55.1 & 27.4 & 65.7 \\
YOLOv3 \cite{redmon2018yolov3} & Darknet-53 & 57.1 & 72.2 & 29.2 & 74.0 & 78.6 & 31.2 & 69.7 & 26.9 & 48.6 & 54.4 & 31.1 & 61.1 & 44.9 & 49.7 & 87.4 & 70.6 & 68.7 & 87.3 & 29.4 & 48.3 & 78.7 \\
\midrule
\multirow{3}{*}{Faster R-CNN \cite{ren2015faster}} & VGG-16 & 54.1 & 53.6 & 49.3 & 78.8 & 66.2 & 28.0 & 70.9 & 62.3 & 69.0 & 55.2 & 68.0 & 56.9 & 50.2 & 50.1 & 27.7 & 73.0 & 39.8 & 75.2 & 38.6 & 23.6 & 45.4 \\
& ResNet-50 & 63.1 & 54.1 & 71.4 & 63.3 & 81.0 & 42.6 & 72.5 & 57.5 & 68.7 & 62.1 & 73.1 & 76.5 & 42.8 & 56.0 & 71.8 & 57.0 & 53.5 & 81.2 & 53.0 & 43.1 & 80.9 \\
& ResNet-101 & 65.1 & 54.0 & 74.5 & 63.3 & 80.7 & 44.8 & 72.5 & 60.0 & 75.6 & 62.3 & 76.0 & 76.8 & 46.4 & 57.2 & 71.8 & 68.3 & 53.8 & 81.1 & 59.5 & 43.1 & 81.2 \\
\midrule
\multirow{2}{*}{RetinaNet \cite{lin2017focal}} & ResNet-50 & 65.7 & 53.7 & 77.3 & 69.0 & 81.3 & 44.1 & 72.3 & 62.5 & 76.2 & 66.0 & 77.7 & 74.2 & 50.7 & 59.6 & 71.2 & 69.3 & 44.8 & 81.3 & 54.2 & 45.1 & 83.4 \\
& ResNet-101 & 66.1 & 53.3 & 77.0 & 69.3 & 85.0 & 44.1 & 73.2 & 62.4 & 78.6 & 62.8 & 78.6 & 76.6 & 49.9 & 59.6 & 71.1 & 68.4 & 45.8 & 81.3 & 55.2 & 44.4 & 85.5 \\
\midrule
\multirow{2}{*}{PANet \cite{liu2018path}} & ResNet-50 & 63.8 & 61.9 & 70.4 & 71.0 & 80.4 & 38.9 & 72.5 & 56.6 & 68.4 & 60.0 & 69.0 & 74.6 & 41.6 & 55.8 & 71.7 & 72.9 & 62.3 & 81.2 & 54.6 & 48.2 & 86.7 \\
& ResNet-101 & 66.1 & 60.2 & 72.0 & 70.6 & 80.5 & 43.6 & 72.3 & 61.4 & 72.1 & 66.7 & 72.0 & 73.4 & 45.3 & 56.9 & 71.7 & 70.4 & 62.0 & 80.9 & 57.0 & 47.2 & 84.5 \\
\midrule
CBD-E \cite{zhang2020contextual} & ResNet-101 & 67.8 & 54.2 & 77.0 & 71.5 & 87.1 & 44.6 & 75.4 & 63.5 & 76.2 & 65.3 & 79.3 & 79.5 & 47.5 & 59.3 & 69.1 & 69.7 & 64.3 & 84.5 & 59.4 & 44.7 & 83.1 \\
CSFF \cite{cheng2020cross} & ResNet-101 & 68.0 & 57.2 & 79.6 & 70.1 & 87.4 & 46.1 & 76.6 & 62.7 & 82.6 & 73.2 & 78.2 & 81.6 & 50.7 & 59.5 & 73.3 & 63.4 & 58.5 & 85.9 & 61.9 & 42.9 & 86.9 \\
FSoD-Net \cite{wang2021fsod} & MSE-Net & 71.8 & 88.9 & 66.9 & 86.8 & 90.2 & 45.5 & 79.6 & 48.2 & 86.9 & 75.5 & 67.0 & 77.3 & 53.6 & 59.7 & 78.3 & 69.9 & 75.0 & 91.4 & 52.3 & 52.0 & 90.6 \\
HawkNet \cite{lin2020novel} & ResNet-50 & 72.0 & 65.7 & 84.2 & 76.1 & 87.4 & 45.3 & 79.0 & 64.5 & 82.8 & 72.4 & 50.2 & 82.5 & \textbf{74.7} & 59.6 & \textbf{89.7} & 66.0 & 70.8 & 87.2 & 61.4 & 52.8 & 88.2 \\
SEPC \cite{wang2020scale} & ResNet-50 & 75.6 & \textbf{92.8} & 81.7 & 94.3 & 83.3 & 46.2 & 91.2 & 69.5 & 68.9 & 59.8 & 76.3 & 76.9 & 63.1 & 63.5 & 86.0 & 82.6 & 82.1 & 90.5 & 55.2 & 70.9 & 77.9 \\
TridentNet \cite{li2019scale} & ResNet-50 & 76.5 & 91.5 & 76.7 & 92.1 & 82.9 & 46.3 & 91.0 & 67.6 & 70.5 & 59.6 & 79.5 & 80.8 & 68.3 & 65.5 & 84.7 & \textbf{90.2} & 80.9 & 89.3 & 65.9 & 68.7 & 78.2 \\
DetectoRS \cite{qiao2021detectors} & ResNet-50 & 76.4 & 91.4 & 78.6 & \textbf{92.6} & 84.2 & 49.0 & \textbf{92.8} & 67.3 & 68.0 & 59.4 & 77.0 & 78.0 & 67.9 & 65.4 & 88.7 & 86.0 & \textbf{81.0} & 89.3 & 62.0 & \textbf{71.3} & 77.8 \\
\midrule
SCRDet++\tnote{\dag} \cite{yang2020scrdet++} & ResNet-50 & 69.4 & 64.3 & 79.0 & 73.2 & 85.7 & 45.8 & 76.0 & 68.4 & 79.3 & 68.9 & 77.7 & 77.9 & 56.7 & 62.2 & 70.4 & 67.7 & 60.4 & 80.9 & 63.7 & 44.4 & 84.6 \\
SCRDet++\tnote{\ddag} & ResNet-50 & 73.2 & 66.4 & 83.4 &74.3 & 87.3 & 52.5 & 78.0 & 70.1 & 84.2 & 78.0 & 80.7 & 81.3 & 56.8 & 63.7 & 73.3 & 71.9 & 71.2 & 83.4 & 62.3 & 55.6 & 90.0 \\
SCRDet++\tnote{\dag} + Aug. & ResNet-101 & 75.1 & 71.9 & 85.0 & 79.5 & 88.9 & 52.3 & 79.1 & 77.6 & 89.5 & 77.8 & 84.2 & 83.1 & 64.2 & 65.6 & 71.3 & 76.5 & 64.5 & 88.0 & 70.9 & 47.1 & 85.1 \\
SCRDet++\tnote{\ddag} + Aug. & ResNet-101 & 77.8 & 80.8 & 87.7 & 80.5 & 89.8 & 57.8 & 80.9 & 75.2 & 90.0 & \textbf{82.9} & 84.5 & 83.6 & 63.2 & 67.3 & 72.6 & 79.2 & 70.4 & 90.0 & 70.7 & 58.8 & 90.3 \\
\midrule
\textbf{CATNet}\tnote{\dag} & ResNet-50 & 74.0 & 70.0 & 86.9 & 76.9 & 89.9 & 48.1 & 82.4 & 73.7 & 86.6 & 69.3 & 84.0 & 81.8 & 60.3 & 63.3 & 84.9 & 70.9 & 61.9 & 88.5 & 63.0 & 48.0 & 90.0 \\
\textbf{CATNet}\tnote{\ddag} & ResNet-50 & 75.8 & 71.3 & 88.9 & 74.3 & 89.5 & 53.2 & 81.1 & 74.3 & 89.1 & 75.4 & 84.8 & 85.1 & 63.5 & 65.0 & 83.3 & 76.3 & 67.9 & 88.9 & 71.8 & 44.8 & 88.1 \\
\textbf{CATNet}\tnote{\dag} + Aug. & ResNet-50 & 78.2 & 84.1 & 90.6 & 84.5 & 91.8 & 51.5 & 83.9 & 79.3 & 92.0 & 70.8 & \textbf{86.6} & 85.2 & 63.4 & 65.8 & 86.7 & 80.4 & 66.2 & 91.6 & 70.6 & 48.7 & 89.6 \\
\textbf{CATNet}\tnote{\ddag} + Aug. & ResNet-50 & \textbf{80.6} & 89.3 & \textbf{92.0} & 84.7 & \textbf{92.2} & \textbf{58.3} & 84.9 & \textbf{80.8} & \textbf{93.2} & 80.2 & 85.8 & \textbf{88.3} & 68.8 & \textbf{68.8} & 84.1 & 82.0 & 70.8 & \textbf{92.2} & \textbf{76.9} & 46.7 & \textbf{91.1} \\
\bottomrule
\end{tabularx}
\vspace{1mm}
\hspace{0.5mm}
\tnote{\dag}\,RetinaNet\,based,\, \tnote{\ddag}\,Faster\,R-CNN\,based,\, Aug.\,-\;Multi-Scale\,Training
\end{threeparttable}
\end{table*}

\begin{table*}
\begin{floatrow}
\renewcommand\tabcolsep{0pt}
\footnotesize
\hspace{-2.2mm}
\ttabbox[0.64\textwidth]{
\caption{Class-wise instance segmentation results on NWPU VHR-10 test set.}}{
\label{tab:4}
\begin{threeparttable}
\begin{tabularx}{\linewidth}{@{\hspace{0.15cm}}p{2.15cm}<{\raggedright}|@{\hspace{0.15cm}}p{1.675cm}<{\raggedright}|p{0.85cm}<{\centering}|@{\hspace{0.05cm}}p{0.65cm}<{\centering}p{0.65cm}<{\centering}p{0.65cm}<{\centering}p{0.65cm}<{\centering}p{0.65cm}<{\centering}p{0.65cm}<{\centering}p{0.65cm}<{\centering}p{0.65cm}<{\centering}p{0.65cm}<{\centering}p{0.65cm}<{\centering}}
\toprule
\textbf{Method}&\textbf{Backbone}&\textbf{AP}\boldmath{$_m$}&\textbf{AI}&\textbf{SH}&\textbf{ST}&\textbf{BD}&\textbf{TC}&\textbf{BC}&\textbf{GT}&\textbf{HA}&\textbf{BR}&\textbf{VE}\\
\midrule
SOLO \cite{wang2019solo} & ResNet-101 & 19.3 & 2.5 & 9.6 & 0.3 & 35.4 & 1.6 & 20.3 & 90.8 & 7.3 & 15.8 & 9.7 \\
RDSNet \cite{wang2019rdsnet} & ResNet-101 & 43.2 & 5.0 & 41.5 & 50.5 & 79.9 & 30.1 & 42.1 & 87.8 & 27.4 & 26.2 & 41.5 \\
CondInst \cite{tian2020conditional} & ResNet-101 & 58.5 & 26.7 & 46.1 & 73.4 & 77.7 & 69.7 & 74.0 & 89.1 & 46.8 & 35.4 & 46.2 \\
\midrule
Mask R-CNN \cite{he2017mask} & ResNet-101 & 58.3 & 28.4 & 52.8 & 69.6 & 81.4 & 59.6 & 69.6 & 84.3 & 60.7 & 25.8 & 50.6 \\
MS R-CNN \cite{huang2019mask} & ResNet-101 & 59.5 & 29.6 & 52.5 & 69.1 & 81.8 & 61.7 & 72.4 & 85.4 & 59.6 & 30.3 & 52.5 \\
HTC \cite{chen2019hybrid} & ResNet-101 & 61.4 & 28.7 & 57.9 & 72.3 & 83.3 & 64.8 & 73.4 & 87.6 & 63.0 & 28.0 & 54.4 \\
SCNet \cite{vu2021scnet} & ResNet-101 & 62.3 & \textbf{54.4} & 58.6 & 70.0 & 85.8 & 69.5 & 72.9 & 89.1 & 64.4 & 24.7 & 55.1 \\
HQ-ISNet \cite{su2020hq} & HRFPN-W40 & 62.7 & 38.7 & 59.2 & 75.1 & 85.8 & 72.2 & 69.2 & 87.9 & 60.2 & 20.9 & 57.8 \\
ARE-Net \cite{zeng2021lightweight} & ResNet-101 & 64.8 & 39.9 & 58.5 & 72.7 & 85.2 & 69.5 & 77.3 & 88.5 & 65.0 & 36.2 & 55.2 \\
CPISNet \cite{zeng2021cpisnet} & AFEN-4GF & 66.1 & 41.5 & 57.6 & 74.2 & 86.2 & 73.3 & 75.7 & 91.6 & \textbf{67.6} & 35.9 & 57.4 \\
\midrule
\textbf{CATNet} & ResNet-50 & 69.1 & 46.0 & 60.7 & 84.2 & 87.9 & 70.3 & 73.2 & 91.5 & 62.9 & 46.9 & 67.0 \\
\textbf{CATNet} + Aug. & ResNet-50 & \textbf{73.3} & 51.9 & \textbf{64.4} & \textbf{87.1} & \textbf{89.4} & \textbf{75.8} & \textbf{79.7} & \textbf{95.0} & 65.0 & \textbf{53.2} & \textbf{72.0} \\
\bottomrule
\end{tabularx}
\vspace{1.3mm}
\hspace{0.5mm}
Aug.\,-\;Multi-Scale\,Training
\end{threeparttable}}
\hspace{-2.9mm}
\ttabbox[0.3475\textwidth]{
\caption{Experimental results on HRSID test set.}}{
\label{tab:5}
\begin{threeparttable}
\begin{tabularx}{\linewidth}{@{\hspace{0.15cm}}p{2.675cm}<{\raggedright}|@{\hspace{0.15cm}}p{1.7cm}<{\raggedright}|@{\hspace{0.08cm}}p{0.7cm}<{\centering}@{\hspace{0.02cm}}p{0.7cm}<{\centering}}
\toprule
\textbf{Method}&\textbf{Backbone}&\textbf{AP}\boldmath{$_b$}&\textbf{AP}\boldmath{$_m$}\\
\midrule
RetinaNet \cite{lin2017focal} & ResNet-101 & 59.8 & -- \\
Faster R-CNN \cite{ren2015faster} & ResNet-101 & 63.9 & -- \\
Cascade R-CNN \cite{cai2018cascade} & ResNet-101 & 66.8 & -- \\
HRDSNet \cite{wei2020hrsid} & HRFPN-W40 & 69.4 & -- \\
\midrule
Mask R-CNN \cite{he2017mask} & ResNet-101 & 65.4 & 54.3 \\
MS R-CNN \cite{huang2019mask} & ResNet-101 & 64.9 & 54.4 \\
Cascade R-CNN \cite{cai2018cascade} & ResNet-101 & 67.6 & 54.7 \\
HTC \cite{chen2019hybrid} & ResNet-101 & 68.4 & 55.4 \\
HQ-ISNet \cite{su2020hq} & HRFPN-W40 & 66.7 & 54.2 \\
GCBANet \cite{ke2022gcbanet} & ResNet-101 & 69.4 & 57.3 \\
\midrule
\textbf{CATNet} & ResNet-50 & 71.7 & 58.2 \\
\textbf{CATNet} + Aug. & ResNet-50 & \textbf{73.3} & \textbf{59.6} \\
\bottomrule
\end{tabularx}
\vspace{1.3mm}
\hspace{0.5mm}
Aug.\,-\;Multi-Scale\,Training
\end{threeparttable}}
\end{floatrow}
\end{table*}

\begin{figure*}
\centering
\includegraphics[width=\textwidth]{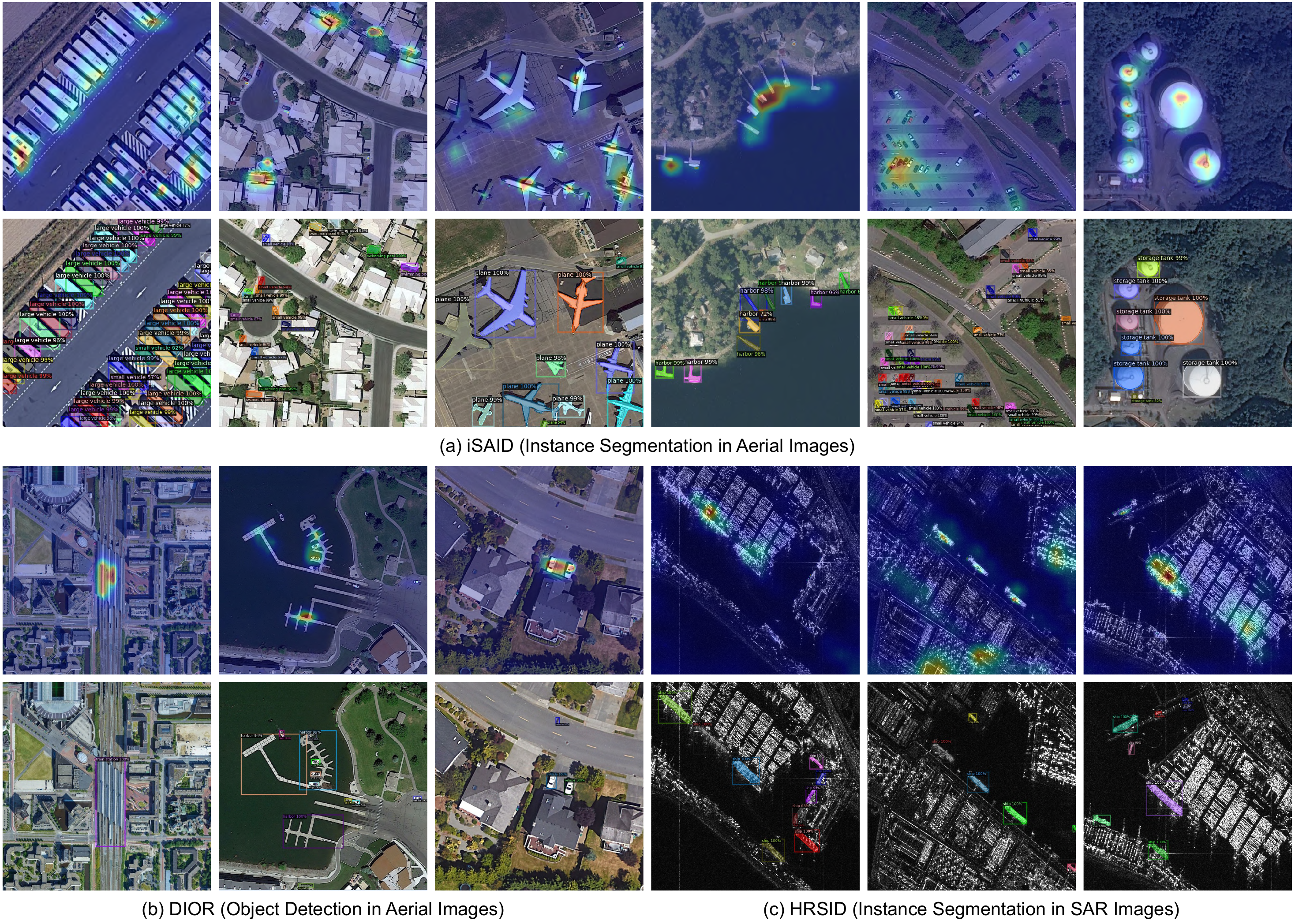}
\caption{Qualitative results on multiple datasets. We visualize the spatial context attention maps in SCP (odd rows) and the final outputs (even rows) on iSAID, DIOR, and HRSID datasets. The results demonstrate that our model has the ability to perform object detection and instance segmentation accurately in both optical remote sensing images and SAR images.}
\label{fig:8}
\end{figure*}

The architecture of SCP is shown in Fig.~\ref{fig:3}~(b). This module also has a pyramid structure, thus can be easily inserted after the backbone or neck. Each layer consists of a context aggregation block (CABlock) with a residual connection. The detailed design of this block is presented in Fig.~\ref{fig:5}~(d). In each block, pixel-wise spatial context is aggregated by
\begin{gather}
Q_i^j = P_i^j + a_i^j \cdot \sum_{j=1}^{N_i} \hspace{1.5mm} [\frac{\exp(w_k P_i^j)}{\sum_{m=1}^{N_i} \exp(w_k P_i^m)} \cdot w_v P_i^j]
\end{gather}
where $P_i$ and $Q_i$ denote the input and output feature maps of level $i$ in the feature pyramid, each contains $N_i$ pixels. $j, m \in \{1, N_i\}$ indicate the indices of each pixel. $w_k$ and $w_v$ are linear projection matrices for projecting the feature maps. In practice, we use $1 \times 1$ convolutions to perform the mapping. The formula above simplifies the widely used self-attention mechanism \cite{vaswani2017attention} by replacing the matrix multiplication between query and key with a linear projection, largely reducing the parameters and computational costs. Beyond GCNet, we apply $a_i$, a re-weighting matrix with the same shape as $P_i$ and $Q_i$, to balance the extent of aggregating global spatial context for each pixel. This matrix can also be generated as simply as a linear projection from $P_i$ with sigmoid activation.
\begin{gather}
a_i^j = \frac{1}{1 + \exp(-w_a P_i^j)}
\end{gather}
Similarly, $j \in \{1, N_i\}$ is the matrix index. Here, the output of sigmoid function shall be regarded as the ratio of information that should be aggregated from global context. We conducted extensive visualizations and experiments on the effectiveness of SCP. Fig.~\ref{fig:7} shows the feature aggregation weights in SCP and GCNet, indicating that SCP tends to aggregate global context into areas with similar semantics, \eg from \textit{small vehicle} to \textit{large vheicle}. Furthermore, we also observe that placing SCP after DenseFPN rather than incorporating CABlock in the backbone leads to better performance while reserving a smaller model size. This dues to the sequential basis of multi-domain global context aggregation. Detailed discussions will be conducted in Section~\ref{sec:4}.

\subsection{Hierarchical Region of Interest Extractor}

Most two-stage object detection and instance segmentation methods lack sufficient attention on the RoI extractor, which may cause severe information loss since only a single scale is considered. The original purpose of this design is to enable large proposals to benefit from low-level features that capture higher localization accuracy, while small proposals can obtain more contextual information due to the larger receptive field of high-level features. We argue that such a hard assignment strategy may not be suitable for all proposals. Recent works \cite{liu2018path,rossi2021novel} also proved that simply computing the sum of RoI features cropped from all layers achieves slightly better performance, indicating that leveraging multi-scale features might alleviate the information loss.

In this work, we further deal with this problem by proposing hierarchical region of interest extractor (HRoIE) to perform task-specific hierarchical feature fusion for each instance. This module is inserted after SCP as displayed in Fig.~\ref{fig:3}~(c). Our hypothesis is that humans can easily perform object detection and segmentation because they focalize their attention on objects in a hierarchical manner. For example, when a person tries to classify an object, he or she would look at the object itself at first. If the object's appearance is not discriminative, the person would gradually look at the surrounding things to gain better information. On the opposite, when segmenting an object at pixel level, a human would look at the whole object to have a comprehensive understanding of its shape, and then zoom in to obtain detailed boundary information for accurate segmentation. We implement the idea above by cropping the features of proposals $R_i$ from all feature pyramid levels in $Q_i$ using RoIAlign \cite{he2017mask}, and utilize several attention blocks to fuse the features hierarchically and adaptively per instance and task. As shown in Fig.~\ref{fig:6}, for each task, the RoI features are initialized from an empty matrix and combined with features from different levels in a hierarchical manner by
\begin{gather}
F_{b/m}^j = F_{b/m}^i + R_i^j \cdot Sigmoid([F_{b/m}^i \hspace{-0.5mm} \parallel \hspace{-0.5mm} R_i^j] \cdot w_i)
\end{gather}
Here, $R_i$ denotes the cropped features at level $i$, $F_{b/m}^i$ and $F_{b/m}^j$ represent the aggregated RoI features at different levels, $w_i$ is the linear projection weight, and $\parallel$ means matrix concatenation at channel dimension. The procedure above computes the pixel-wise attention weights for feature aggregation, thus RoI features can be generated adaptively per instance and task. In practice, we adopt a bottom-up path for object detection head and a top-down path for mask prediction head.

\begin{table*}
\caption{Class-wise instance segmentation results on iSAID val set. All the models use ResNet-50 as backbone.}
\label{tab:6}
\renewcommand\tabcolsep{0pt}
\footnotesize
\begin{threeparttable}
\begin{tabularx}{\linewidth}{@{\hspace{0.15cm}}p{2.93cm}<{\raggedright}|p{1.1cm}<{\centering}@{\hspace{-0.15cm}}p{1.1cm}<{\centering}|@{\hspace{0.1cm}}p{0.86cm}<{\centering}p{0.86cm}<{\centering}p{0.86cm}<{\centering}p{0.86cm}<{\centering}p{0.86cm}<{\centering}p{0.86cm}<{\centering}p{0.86cm}<{\centering}p{0.86cm}<{\centering}p{0.86cm}<{\centering}p{0.86cm}<{\centering}p{0.86cm}<{\centering}p{0.86cm}<{\centering}p{0.86cm}<{\centering}p{0.86cm}<{\centering}p{0.86cm}<{\centering}}
\toprule
\textbf{Method}&\textbf{AP}\boldmath{$_b$}&\textbf{AP}\boldmath{$_m$}&\textbf{SH}&\textbf{ST}&\textbf{BD}&\textbf{TC}&\textbf{BC}&\textbf{GT}&\textbf{BR}&\textbf{LV}&\textbf{SV}&\textbf{HE}&\textbf{SP}&\textbf{RO}&\textbf{SB}&\textbf{PL}&\textbf{HA}\\
\midrule
\textbf{Mask R-CNN} \cite{he2017mask} & 43.8 & 36.5 & 47.2 & 36.8 & 54.0 & 76.7 & 32.3 & 32.7 & 22.9 & 39.4 & 15.3 & 4.9 & 30.1 & 37.3 & 46.4 & 48.1 & 26.8 \\
+ DenseFPN & 45.5 & 37.7 & \textbf{48.8} & 37.7 & 55.6 & 77.5 & 35.8 & 33.4 & 23.8 & 39.6 & 16.3 & 5.7 & 29.9 & \textbf{39.8} & 46.7 & 49.2 & 28.4 \\
+ SCP & 45.9 & 37.8 & 48.5 & 38.2 & \textbf{55.8} & \textbf{77.8} & 38.6 & 32.5 & 25.0 & \textbf{39.8} & 16.2 & 5.9 & 28.9 & 39.3 & 46.4 & 49.6 & 27.9 \\
+ HRoIE & \textbf{46.2} & \textbf{38.5} & 48.4 & \textbf{38.4} & 55.4 & \textbf{77.8} & \textbf{40.4} & \textbf{35.3} & \textbf{25.8} & 39.5 & \textbf{16.4} & \textbf{6.3} & \textbf{31.0} & 38.3 & \textbf{47.8} & \textbf{50.2} & \textbf{29.0} \\
\midrule
\textbf{Cascade R-CNN} \cite{cai2018cascade} & 45.8 & 37.3 & 47.7 & 37.0 & 55.6 & 77.4 & 35.2 & 33.7 & 22.6 & 40.2 & 15.9 & 5.9 & 29.8 & 38.6 & 47.2 & 47.5 & 27.6 \\
+ DenseFPN & 46.8 & 38.2 & \textbf{49.6} & 38.9 & 55.1 & \textbf{78.8} & 34.4 & 34.4 & 25.1 & \textbf{41.4} & \textbf{16.8} & 4.8 & \textbf{31.2} & 41.0 & 45.9 & 49.9 & 28.4 \\
+ SCP & 47.4 & 38.4 & 49.5 & 39.0 & \textbf{55.9} & 77.9 & 35.6 & 34.8 & 25.3 & 40.7 & 16.7 & 5.7 & 31.1 & 41.2 & \textbf{47.5} & 49.7 & 28.8 \\
+ HRoIE & \textbf{47.8} & \textbf{38.9} & 49.1 & \textbf{39.2} & 55.7 & 78.3 & \textbf{37.4} & \textbf{35.5} & \textbf{25.4} & 41.2 & \textbf{16.8} & \textbf{8.7} & \textbf{31.2} & \textbf{41.8} & 47.1 & \textbf{50.0} & \textbf{29.9} \\
\midrule
\textbf{MS R-CNN} \cite{huang2019mask} & 44.0 & 37.8 & 48.6 & 36.6 & 54.9 & 77.0 & 36.9 & 35.2 & 22.5 & 39.9 & 15.5 & 9.7 & 30.1 & 39.8 & 44.7 & 49.2 & 29.7 \\
+ DenseFPN & 45.4 & 38.5 & 49.4 & 37.5 & 55.3 & 77.4 & 39.2 & 35.4 & 24.8 & \textbf{41.3} & 16.3 & 6.9 & 31.4 & 37.6 & 47.1 & 50.6 & 29.5 \\
+ SCP & 45.9 & 38.9 & 49.2 & \textbf{37.9} & 56.2 & 77.8 & \textbf{40.8} & 35.4 & 24.5 & 40.9 & 16.2 & 8.6 & 31.8 & 40.5 & 46.6 & 50.3 & 29.7 \\
+ HRoIE & \textbf{46.1} & \textbf{39.3} & \textbf{49.5} & 37.8 & \textbf{56.6} & \textbf{77.9} & 37.5 & \textbf{37.7} & \textbf{25.0} & 41.2 & \textbf{16.4} & \textbf{9.8} & \textbf{32.0} & \textbf{40.6} & \textbf{50.1} & \textbf{51.0} & \textbf{30.2} \\
\midrule
\textbf{PointRend} \cite{kirillov2019pointrend} & 43.4 & 37.4 & 47.8 & 36.6 & 55.6 & 76.8 & 34.3 & 33.3 & 23.0 & 41.2 & 15.8 & 5.0 & 29.9 & 38.0 & 47.4 & 49.8 & 29.8 \\
+ DenseFPN & 44.6 & 38.3 & 49.4 & 39.2 & 54.7 & 77.6 & 35.7 & 33.4 & 23.8 & 41.7 & 17.0 & 5.5 & \textbf{30.9} & 37.7 & 47.6 & 52.0 & \textbf{31.7} \\
+ SCP & 45.2 & 38.7 & 50.9 & 39.4 & 55.7 & 78.4 & 36.6 & 33.7 & 25.0 & \textbf{43.1} & \textbf{16.8} & 5.0 & 30.1 & 39.2 & 46.2 & 52.7 & 31.4 \\
+ HRoIE & \textbf{45.7} & \textbf{39.3} & \textbf{51.6} & \textbf{39.7} & \textbf{55.8} & \textbf{78.6} & \textbf{37.7} & \textbf{36.0} & \textbf{25.6} & 42.8 & \textbf{16.8} & \textbf{5.7} & 29.8 & \textbf{39.9} & \textbf{47.7} & \textbf{53.1} & 31.6 \\
\bottomrule
\end{tabularx}
\vspace{1.3mm}
\hspace{0.5mm}
AP$_b$\,-\,Bounding box\,AP,\, AP$_m$\,-\,Mask\,AP
\end{threeparttable}
\end{table*}

\section{Experiments}\label{sec:4}

In this section, we extensively evaluate the proposed method on iSAID, DIOR, NWPU VHR-10, and HRSID datasets, in which the first three datasets are based on optical remote sensing images, while the last one is for synthetic aperture radar (SAR) images. Note that DIOR only provides bounding box annotations, thus we only evaluate the object detection performance on this dataset.

\begin{table*}
\renewcommand\tabcolsep{0pt}
\caption{Detailed comparisons on iSAID val set. The baseline model is Mask R-CNN with ResNet-50-FPN as backbone and neck.}
\label{tab:7}
\footnotesize
\hspace{-2mm}\subfloat[Types of Multi-Scale Feature Propagation Modules]{
\begin{threeparttable}
\begin{tabularx}{0.488\linewidth}{@{\hspace{0.15cm}}p{2.35cm}<{\raggedright}|p{1cm}<{\centering}|p{1.1cm}<{\centering}|@{\hspace{0.1cm}}p{0.75cm}<{\centering}p{0.75cm}<{\centering}@{\hspace{0.08cm}}|p{1.3cm}<{\centering}|p{1.2cm}<{\centering}}
\toprule
\textbf{Method}&\textbf{Chn.}&\textbf{Depth}&\textbf{AP}\boldmath{$_b$}&\textbf{AP}\boldmath{$_m$}&\textbf{\#Params}&\textbf{FLOPs}\\
\midrule
Baseline & 256 & -- & 43.8 & 36.5 & 44.05M & 114.96G \\
PAFPN \cite{liu2018path} & 256 & -- & 44.4 & 37.0 & 47.59M & 121.31G \\
HRFPN \cite{wang2020deep} & 256 & -- & 44.1 & 36.8 & 44.64M & 129.09G \\
CARAFE \cite{wang2021carafe++} & 256 & -- & 44.1 & 36.9 & 49.65M & 115.71G \\
BFP \cite{pang2019libra} & 256 & -- & 44.2 & 37.0 & 44.31M & 115.23G \\
AugFPN \cite{guo2020augfpn} & 256 & -- & 44.3 & 36.8 & 45.82M & 115.03G \\
TridentNet \cite{li2019scale} & 256 & -- & 42.6 & 32.9 & 35.28M & 838.99G \\
RFP \cite{qiao2021detectors} & 256 & -- & 45.3 & 37.4 & 68.80M & 153.88G \\
BiFPN \cite{tan2019efficientdet} & 256 & -- & 44.1 & 36.8 & 45.89M & 112.49G \\
\midrule
\multirow{4}{*}{NAS-FPN \cite{ghiasi2019nas}} & \multirow{4}{*}{256} & 1 & 44.2 & 36.9 & 45.89M & 119.86G \\
&& 3 & 45.3 & 37.4 & 54.16M & 155.30G \\
&& 5 & 45.5 & 37.6 & 62.42M & 190.74G \\
&& 7 & 45.6 & 37.7 & 70.69M & 226.18G \\
\midrule
\multirow{4}{*}{FPG \cite{chen2020feature}} & \multirow{4}{*}{128} & 5 & 43.7 & 36.3 & 38.96M & 92.75G \\
&& 7 & 44.2 & 36.6 & 41.66M & 102.23G \\
&& 9 & 44.5 & 36.9 & 44.36M & 111.70G \\
&& 11 & 44.6 & 37.1 & 47.05M & 121.18G \\
\midrule
\multirow{4}{*}{FPG \cite{chen2020feature}} & \multirow{4}{*}{256} & 5 & 44.2 & 36.8 & 60.79M & 143.19G \\
&& 7 & 44.8 & 37.1 & 71.55M & 181.04G \\
&& 9 & 45.1 & 37.3 & 82.32M & 218.90G \\
&& 11 & 45.2 & 37.4 & 93.09M & 256.75G \\
\midrule
\multirow{4}{*}{\textbf{DenseFPN} (ours)} & \multirow{4}{*}{128} & 1 & 44.0 & 36.5 & 34.87M & 84.81G \\
&& 3 & 44.4 & 36.8 & 35.94M & 89.46G \\
&& 5 & 44.7 & 37.0 & 37.02M & 94.10G \\
&& 7 & 44.9 & 37.1 & 38.09M & 98.75G \\
\midrule
\multirow{4}{*}{\textbf{DenseFPN} (ours)} & \multirow{4}{*}{256} & 1 & 44.7 & 37.1 & 44.42M & 111.45G \\
&& 3 & 45.2 & 37.5 & 48.70M & 130.05G \\
&& 5 & 45.5 & 37.7 & 52.98M & 148.64G \\
&& 7 & \textbf{45.7} & \textbf{37.8} & 57.26M & 167.23G \\
\bottomrule
\end{tabularx}
\vspace{1.5mm}
\hspace{0.5mm}
Chn.\,-\,Feature\,Channels
\end{threeparttable}}
\hspace{2.3mm}
\subfloat[Types of Spatial Context Modules]{
\begin{threeparttable}
\begin{tabularx}{0.488\linewidth}{@{\hspace{0.15cm}}p{2.075cm}<{\raggedright}|p{1.425cm}<{\centering}|p{0.95cm}<{\centering}|@{\hspace{0.1cm}}p{0.75cm}<{\centering}p{0.75cm}<{\centering}@{\hspace{0.08cm}}|p{1.3cm}<{\centering}|p{1.2cm}<{\centering}}
\toprule
\textbf{Method}&\textbf{Position}&\textbf{Red.}&\textbf{AP}\boldmath{$_b$}&\textbf{AP}\boldmath{$_m$}&\textbf{\#Params}&\textbf{FLOPs}\\
\midrule
Baseline & -- & -- & 43.8 & 36.5 & 44.05M & 114.96G \\
GABlock \cite{zhu2019empirical} & C$_3$\;$\sim$\;C$_5$ & -- & 44.5 & 37.0 & 143.25M & 156.52G \\
CCBlock \cite{huang2019ccnet} & C$_3$\;$\sim$\;C$_5$ & -- & 44.1 & 36.7 & 68.97M & 132.43G \\
\midrule
\multirow{4}{*}{NLBlock \cite{wang2018non}} & \multirow{4}{*}{C$_3$\;$\sim$\;C$_5$} & 2 & 44.2 & 37.0 & 83.93M & 142.92G \\
&& 4 & 44.0 & 36.8 & 63.99M & 128.95G \\
&& 8 & 43.8 & 36.7 & 54.03M & 121.97G \\
&& 16 & 43.5 & 36.5 & 49.04M & 118.47G \\
\midrule
\multirow{4}{*}{GCBlock \cite{cao2023global}} & \multirow{4}{*}{C$_3$\;$\sim$\;C$_5$} & 2 & 44.5 & 37.2 & 64.02M & 115.00G \\
&& 4 & 44.3 & 37.1 & 54.05M & 114.99G \\
&& 8 & 44.1 & 36.9 & 49.06M & 114.99G \\
&& 16 & 44.0 & 36.8 & 46.57M & 114.98G \\
\midrule
\multirow{4}{*}{\textbf{CABlock} (ours)} & \multirow{4}{*}{C$_3$\;$\sim$\;C$_5$} & 2 & 44.5 & 37.6 & 64.02M & 121.99G \\
&& 4 & 44.4 & 37.4 & 54.05M & 118.50G \\
&& 8 & 44.3 & 37.2 & 49.07M & 116.75G \\
&& 16 & 44.2 & 37.1 & 46.58M & 115.87G \\
\midrule
GABlock \cite{zhu2019empirical} & P$_2$\;$\sim$\;P$_6$ & -- & 43.9 & 36.6 & 45.66M & 119.20G \\
CCBlock \cite{huang2019ccnet} & P$_2$\;$\sim$\;P$_6$ & -- & 43.6 & 36.2 & 44.46M & 116.76G \\
NLBlock \cite{wang2018non} & P$_2$\;$\sim$\;P$_6$ & 1 & 44.0 & 36.9 & 45.36M & 120.71G \\
GCBlock \cite{cao2023global} & P$_2$\;$\sim$\;P$_6$ & 1 & 44.2 & 37.2 & 44.71M & 114.97G \\
\midrule
\textbf{SCP} (ours) & P$_2$\;$\sim$\;P$_6$ & 1 & \textbf{44.6} & \textbf{37.7} & 44.71M & 116.41G \\
\bottomrule
\end{tabularx}
\vspace{1.5mm}
\hspace{0.5mm}
Red.\,-\;Channel\,Reduction\,Rate
\end{threeparttable}}
\vspace{-2.825cm}

\hspace{9.1cm}
\subfloat[Orders of the Proposed Modules]{
\begin{threeparttable}
\begin{tabularx}{0.488\linewidth}{@{\hspace{0.15cm}}p{4.5cm}<{\raggedright}|@{\hspace{0.1cm}}p{0.75cm}<{\centering}p{0.75cm}<{\centering}@{\hspace{0.08cm}}|p{1.3cm}<{\centering}|p{1.2cm}<{\centering}}
\toprule
\textbf{Module Order}&\textbf{AP}\boldmath{$_b$}&\textbf{AP}\boldmath{$_m$}&\textbf{\#Params}&\textbf{FLOPs}\\
\midrule
Backbone (\textit{w.} CABlock) $\rightarrow$ DFPN & 45.4 & 37.5 & 72.95M & 155.67G \\
Backbone $\rightarrow$ SCP $\rightarrow$ DFPN & 45.5 & 37.6 & 53.64M & 150.09G \\
Backbone $\rightarrow$ DFPN $\rightarrow$ SCP & \textbf{45.9} & \textbf{37.8} & 53.64M & 150.09G \\
\bottomrule
\end{tabularx}
\vspace{1.5mm}
\hspace{0.5mm}
DFPN\,-\,DenseFPN
\end{threeparttable}}

\hspace{-2.4mm}
\subfloat[Types of Region of Interest Extractors]{
\begin{threeparttable}
\begin{tabularx}{0.435\linewidth}{@{\hspace{0.15cm}}p{1.95cm}<{\raggedright}|p{1.5cm}<{\centering}|@{\hspace{0.1cm}}p{0.75cm}<{\centering}p{0.75cm}<{\centering}@{\hspace{0.08cm}}|p{1.3cm}<{\centering}|p{1.2cm}<{\centering}}
\toprule
\textbf{Method}&\textbf{Direction}&\textbf{AP}\boldmath{$_b$}&\textbf{AP}\boldmath{$_m$}&\textbf{\#Params}&\textbf{FLOPs}\\
\midrule
Baseline & -- & 43.8 & 36.5 & 44.05M & 114.96G \\
GRoIE \cite{rossi2021novel} & -- & 44.0 & 36.9 & 47.84M & 574.92G \\
\midrule
SUM & -- & 43.9 & 36.7 & 44.05M & 114.96G \\
CONCAT & -- & 44.0 & 36.7 & 84.35M & 188.18G \\
ATTENTION & -- & 43.9 & 36.8 & 46.15M & 186.97G \\
\midrule
\textbf{HRoIE} (ours) & $\downarrow$\, +\, $\downarrow$ & 44.2 & 37.1 & 45.10M & 151.00G \\
\textbf{HRoIE} (ours) & $\downarrow$\, +\, $\uparrow$ & 44.1 & 37.0 & 45.10M & 151.00G \\
\textbf{HRoIE} (ours) & $\uparrow$\, +\, $\uparrow$ & 44.3 & 36.9 & 45.10M & 151.00G \\
\textbf{HRoIE} (ours) & $\uparrow$\, +\, $\downarrow$ & \textbf{44.4} & \textbf{37.2} & 45.10M & 151.00G \\
\bottomrule
\end{tabularx}
\vspace{1.5mm}
\hspace{0.5mm}
$\downarrow$\,-\,Top-Down,\, $\uparrow$\,-\,Bottom-Up
\end{threeparttable}}
\hspace{3.6mm}\subfloat[Ablation Study and Efficiency Comparisons]{
\begin{threeparttable}
\begin{tabularx}{0.5425\linewidth}{@{\hspace{0.025cm}}p{1.15cm}<{\centering}p{1.15cm}<{\centering}p{1.15cm}<{\centering}p{1.15cm}<{\centering}|@{\hspace{0.1cm}}p{0.75cm}<{\centering}p{0.75cm}<{\centering}@{\hspace{0.08cm}}|p{1.3cm}<{\centering}|p{1.2cm}<{\centering}|p{1cm}<{\centering}}
\toprule
\textbf{DFPN}&\textbf{SCP}&\textbf{HRoIE}&\textbf{Lite}&\textbf{AP}\boldmath{$_b$}&\textbf{AP}\boldmath{$_m$}&\textbf{\#Params}&\textbf{FLOPs}&\textbf{FPS}\\
\midrule
& & & & 43.8 & 36.5 & 44.05M & 114.96G & 32.69 \\
$\checkmark$ & & & & 45.5 & 37.7 & 52.98M & 148.64G & 29.79 \\
& $\checkmark$ & & & 44.6 & 37.7 & 44.71M & 116.41G & 31.48 \\
& & $\checkmark$ & & 44.4 & 37.2 & 45.10M & 151.00G & 28.83 \\
\midrule
$\checkmark$ & $\checkmark$ & & & 45.9 & 37.8 & 53.64M & 150.09G & 29.06 \\
$\checkmark$ & & $\checkmark$ & & 46.0 & 38.2 & 54.03M & 184.68G & 25.36 \\
& $\checkmark$ & $\checkmark$ & & 44.7 & 38.0 & 45.76M & 152.45G & 28.51 \\
\midrule
$\checkmark$ & $\checkmark$ & $\checkmark$ & & \textbf{46.2} & \textbf{38.5} & 54.69M & 186.12G & 24.05 \\
$\checkmark$ & $\checkmark$ & $\checkmark$ & $\checkmark$ & 44.7 & 37.5 & 37.45M & 103.50G & 35.78 \\
\bottomrule
\end{tabularx}
\vspace{1.5mm}
\hspace{0.5mm}
Lite\,-\,Lite\,version\,(\textit{w.}\,1-layer\,DenseFPN@128)
\end{threeparttable}}
\end{table*}

\subsection{Datasets and Evaluation Metrics}

\vspace{0.5mm}\noindent\textbullet\;\;\textbf{iSAID \cite{waqas2019isaid}} is a large-scale dataset for instance segmentation in aerial images. The images in iSAID are inherited from DOTA \cite{xia2018dota}, which is popular for oriented object detection. It contains $15$ classes of $655,451$ instances in $2,806$ images, with all the objects independently annotated from scratch. The spatial resolutions of images are in a large range between $800$ and $13,000$. Following previous works \cite{chen2021db,garg2021isdnet,zeng2021cpisnet}, we split them into $800 \times 800$ patches with a stride of $200$ for fair benchmarking with existing methods. When conducting detailed comparisons and ablation studies, we adopt a smaller patch size of $512 \times 512$ with $128$ stride to reduce the training cost. The abbreviations of classes are SH\;-\;ship, ST\;-\;storage tank, BD\;-\;baseball diamond, TC\;-\;tennis court, BC\;-\;basketball court, GT\;-\;ground track field, BR\;-\;bridge, LV\;-\;large vehicle, SV\;-\;small vehicle, HE\;-\;helicopter, SP\;-\;swimming pool, RO\;-\;roundabout, SB\;-\;soccer ball field, PL\;-\;plane, and HA\;-\;harbor.

\vspace{0.5mm}\noindent\textbullet\;\;\textbf{DIOR \cite{li2020object}} is a complex aerial images dataset labeled by horizontal and oriented bounding boxes. It contains $23,463$ images with $190,288$ instances, covering $20$ object classes. Object sizes in DIOR have severe inter-class and intra-class variabilities. The complexity of this dataset is also reflected in different imaging qualities, weathers, and seasons. The abbreviations of classes are AL\;-\;airplane, AR\;-\;airport, BF\;-\;baseball field, BC\;-\;basketball court, BR\;-\;bridge, CH\;-\;chimney, DA\;-\;dam, ES\;-\;expressway service area, ET\;-\;expressway toll station, GC\;-\;golf field, GT\;-\;ground track field, HA\;-\;harbor, OV\;-\;overpass, SH\;-\;ship, ST\;-\;stadium, SA\;-\;storage tank, TC\;-\;tennis court, TS\;-\;train station, VE\;-\;vehicle, and WM\;-\;wind mill.

\vspace{0.5mm}\noindent\textbullet\;\;\textbf{NWPU VHR-10 \cite{cheng2014multi}} was originally designed for object detection in aerial images, and has been enriched with instance-level mask annotations \cite{su2020hq}. It contains $10$ object categories in $800$ high-resolution images, among which $650$ are positive and $150$ are negative without any objects of interest. Since this dataset has no official train/test split available, we randomly select $70\%$ of the images for training while the others for testing. Our splits will be released for fair comparison with the following methods. The abbreviations of classes are AI\;-\;airplane, SH\;-\;ship, ST\;-\;storage tank, BD\;-\;baseball diamond, TC\;-\;tennis court, BC\;-\;basketball court, GT\;-\;ground track field, HA\;-\;harbor, BR\;-\;bridge, and VE\;-\;vehicle.

\vspace{0.5mm}\noindent\textbullet\;\;\textbf{HRSID \cite{wei2020hrsid}} is a recently introduced dataset for ship detection and segmentation in synthetic aperture radar (SAR) images. This dataset contains a total of 5,604 high-resolution SAR images with 16,951 ship instances. All the instances in this dataset are annotated with pixel-level masks. Spatial resolutions of the images are $0.5m$, $1m$, and $3m$.

\vspace{0.5mm}

We follow the standard evaluation metric that utilizes mean average precision (mAP) to measure the detection and segmentation performances. A prediction is considered a true positive (TP) when the bounding box or mask of the object has an intersection over union (IoU) with its corresponding ground truth greater than a threshold $\theta_{IoU}$, and the predicted class label is correct. For iSAID, NWPU VHR-10, and HRSID datasets, we compute the mean of mAPs under $\theta_{IoU}$ ranging from $0.05$ to $0.95$. For DIOR dataset, only the mAPs under $\theta_{IoU} = 0.5$ are considered following the original paper. Aside from the accuracy metrics, we also evaluate the efficiencies of the models using the number of parameters (\#Params) and floating-point operations (FLOPs), measuring the storage and computational efficiencies, respectively. We consider these two metrics since they are both hardware-irrelevant and have been widely adopted in previous studies.

\begin{figure*}
\centering
\vspace{2mm}
\includegraphics[width=\textwidth]{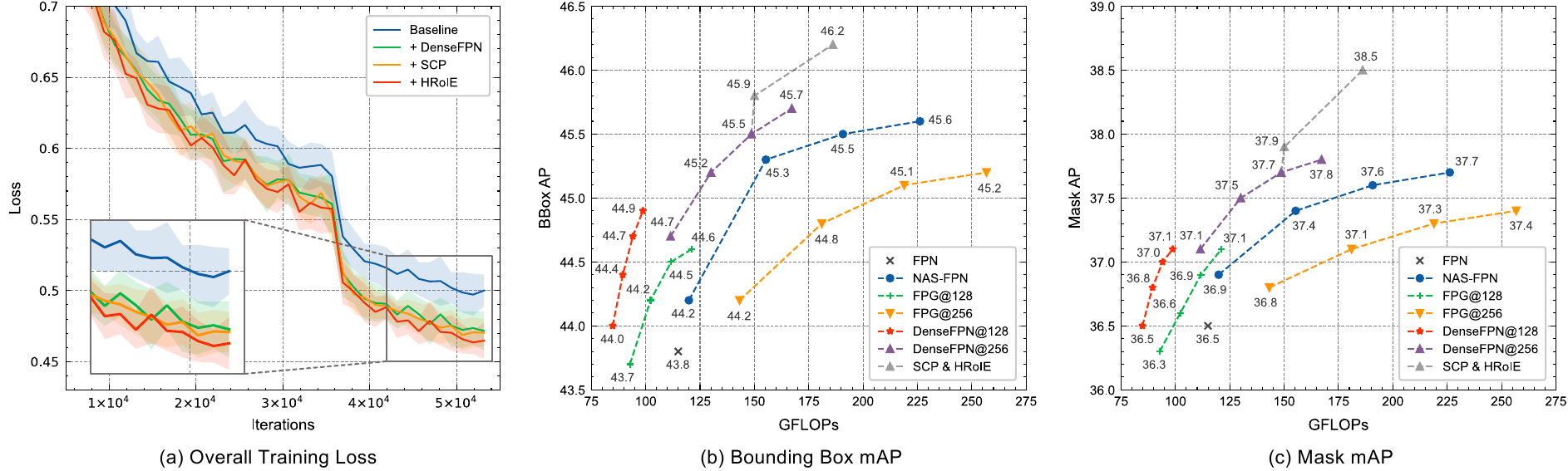}
\caption{Visualizations of training and test metrics on iSAID val set. a) displays the overall training losses with and without our proposed modules. b) and c) shows the performance comparison among different multi-scale propagation modules. Performances of each module are reported with 4 different depths.}
\label{fig:9}
\end{figure*}

\subsection{Implementation Details}

We choose Mask R-CNN \cite{he2017mask}, Faster R-CNN \cite{ren2015faster}, and RetinaNet \cite{lin2017focal} with ResNet-50 \cite{he2016deep} backbone as our baselines for different scenarios. The backbone is pre-trained on ImageNet \cite{krizhevsky2012imagenet} and fine-tuned with the detectors. All the parameters in the first stage are frozen after pre-training. If not specified, $5$ basic blocks are included in all DenseFPN modules. In order to stabilize the training process, synchronized batch normalizations (SyncBN) \cite{ioffe2015batch} are used among intermediate layers. When testing, we adopt Soft-NMS \cite{bodla2017soft} to suppress the duplicate results with IoU larger than $0.5$ and a maximum of $1,000$ predictions would be made for each image, since most objects are heavily overlapped in remote sensing images.

We use stochastic gradient descent (SGD) optimizer with initial learning rate $0.01$, momentum $0.9$, and weight decay $0.0001$ to learn the parameters for all models. Each training batch contains 8 images. On iSAID dataset, we follow the standard $1 \times$ training schedule that drops the learning rate by $0.1$ at epoch $8$ and $11$, and stop training at epoch $12$ for efficient parameter tuning and ablation studies on validation set. When benchmarking on test set, we train our model on both training and validation sets under a $3 \times$ schedule, which is a common setting in existing works \cite{waqas2019isaid,chen2021db}. On DIOR, NWPU VHR-10, and HRSID datasets, we adopt the $3 \times$, $6 \times$, and $3 \times$ training schedules, respectively. All the models are trained with PyTorch \cite{paszke2017automatic} on a compute node with 2 Intel Xeon Platinum 8358 (2.6GHz) Processors, 512GB RAM, and 4 NVIDIA A100 (80G) GPUs. The training of our model on iSAID, DIOR, NWPU VHR-10, and HRSID datasets costs about $27$, $8$, $1$, and $5$ hours, respectively.

\subsection{Instance Segmentation on iSAID Dataset}

We first evaluate our approach on optical remote sensing images. Table~\ref{tab:1} shows the benchmark of instance segmentation methods on iSAID test set. The results are obtained by submitting the predictions onto the official evaluation server, which only accepts one submission per day, avoiding potential model tuning on the test set. Here, $AP^{50}$ and $AP^{75}$ indicate average precisions under $\theta_{IoU} = 0.5$ and $0.75$, while $AP^S$, $AP^M$, and $AP^L$ represent the average precisions for small ($10$\,$\sim$\,$144$ px), medium ($144$\,$\sim$\,$1024$ px), and large (more than $1024$ px) objects, respectively. The evaluation results of CATNet are obtained with two standard settings, with input images re-scaled to $(1400, 800)$ only and with augmentations on the short side at five scales $(1200, 1000, 800, 600, 400)$. Among existing methods, our proposed CATNet achieves state-of-the-art performances under both object detection and instance segmentation metrics, even with a $40\%$ lighter backbone. When adopting multi-scale training, which is a common setting in previous works, our model can be further boosted significantly, demonstrating the scalability of our approach.

To justify the effectiveness of our method, class-wise instance segmentation results are also provided in Table~\ref{tab:2}. Our model obtains better performances on most categories, including ``storage tank'', ``baseball diamond'', and ``tennis court''. On other categories, our method is still comparable with most existing works.

\subsection{Object Detection on DIOR Dataset}

We then evaluate our approach on DIOR dataset. Note that this dataset only provides bounding box annotations for objects, we only validate the object detection performance on it. Such a comparison can also demonstrate the generalization ability of the modules -- they could potentially benefit all the dense prediction tasks including object detection and instance segmentation. Table~\ref{tab:3} presents the class-wise performance comparison with existing methods. Our methods are built upon two meta algorithms for one-stage and two-stage object detection, which are RetinaNet \cite{lin2017focal} and Faster R-CNN \cite{ren2015faster}. Note that one-stage models do not need RoI extractors, so that CATNet$^\dag$ only contains DenseFPN and SCP. The comparison shows that our proposed method outperforms all the previous methods, with or without data augmentations. Moreover, the performances of our model with ResNet-50 backbones are even better than the previous state-of-the-art with ResNet-101 by a noticeable margin. This demonstrates the effectiveness and generalization ability of the proposed modules.

\subsection{Instance Segmentation on NWPU VHR-10 Dataset}

To validate the significance of our method, we also conduct instance segmentation experiments on an extra NWPU VHR-10 Dataset, and the performance comparisons are shown in Table~\ref{tab:4}. The first group includes one-stage methods while the second group contains two-stage methods. Following previous works, we only report the average precisions for mask predictions to focus on the instance segmentation task. The proposed CATNet can still perform better than all the previous approaches with heavier backbones.

\subsection{Instance Segmentation on HRSID Dataset}

Beyond optical remote sensing images, we also evaluate our model on the more challenging SAR images on HRSID dataset. The results are presented in Table~\ref{tab:5}, in which the first group contains object detection models while the second group includes instance segmentation models. Practically, SAR images are considered as single-channel grayscale images, so that each input for the model is constructed by repeating the SAR images for three times along the channel dimension. The experimental results prove that compared with strong baselines for natural or optical remote sensing images, our method significantly works better on both object detection and instance segmentation.

\subsection{Visualizations}

We also provide some visualizations to conduct an in-depth study on the significance and effectiveness of our method.

\vspace{0.5mm}\textbf{Context Aggregation Weights:} To demonstrate the effectiveness and significance of SCP, we visualize the context aggregation weights in GCNet \cite{cao2023global} and SCP in Fig.~\ref{fig:7}. Each row represents the weights of all classes that be aggregated into each class. The sizes and color depths of circles denote the weights in GCNet and SCP, respectively. Larger circles or darker colors indicate higher weights. From the visualization, we can observe that our method tends to aggregate global spatial context from objects with the same category. Some similar (\eg \textit{plane} and \textit{helicopter}) or semantically related (\eg \textit{ship} and \textit{harbor}) categories can also help each other during training. For example, SCP tends to aggregate the information of \textit{small vehicle} to \textit{large vehicle}, since they are semantically similar objects, while GCNet simple fuse \textit{plane} into \textit{large vehicle} as global context is often dominated by large objects. Compared with our method, GCNet does not focus much on similar or semantically related objects, leading to information confounding when aggregating features.

\begin{figure}
\centering
\vspace{2mm}
\includegraphics[width=\linewidth]{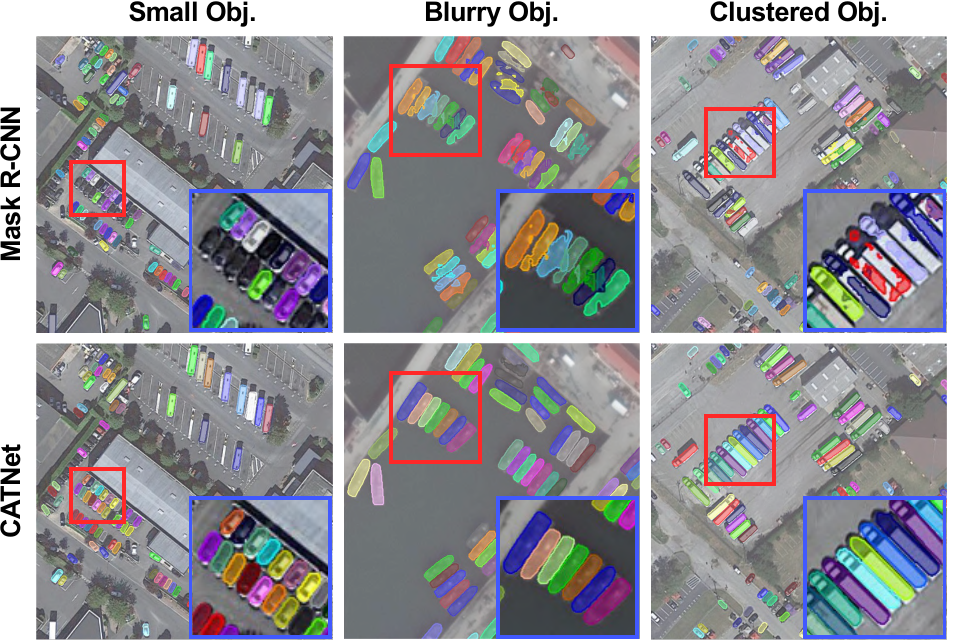}
\caption{Detailed comparisons on mask predictions for small, blurry, and clustered objects. The image patches are from iSAID val set.}
\label{fig:10}
\end{figure}

\vspace{0.5mm}\textbf{Model Predictions and Attention Maps:} Fig.~\ref{fig:8} shows some qualitative results on iSAID, DIOR, and HRSID datasets. Each image patch is visualized by its feature aggregation weights in SCP (top) and the final object detection or instance segmentation results (bottom). The feature aggregation weights indicate that our method focuses more on areas that contain objects rather than pure background, and aggregates them into regions with poor features. Furthermore, we also visualize the detailed comparisons on mask prediction for hard cases in Fig.~\ref{fig:10}. Our model can better handle small, blurry, and clustered objects, which are common in both optical and other kins of remote sensing images. These results demonstrate that our method can effectively detect and segment objects accurately in multiple scenes.

\begin{figure}
\centering
\vspace{2mm}
\includegraphics[width=\linewidth]{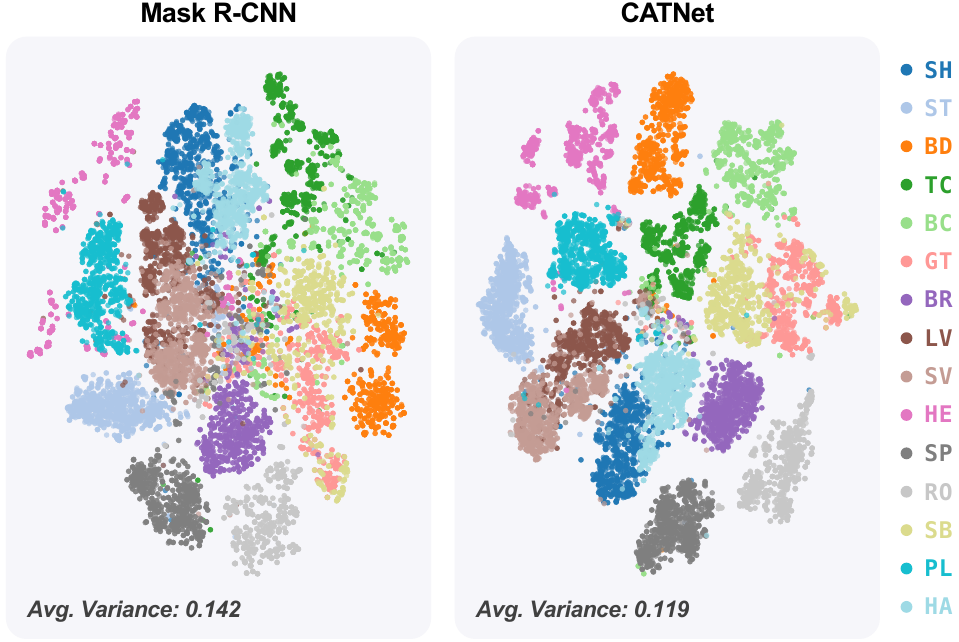}
\caption{Visualization of the sampled RoI features from different models using t-SNE. The averaged class-wise variances are also reported.}
\label{fig:11}
\end{figure}

\vspace{0.5mm}\textbf{Training Process:} In order to study the model training process, we also visualize the overall training losses in Fig.~\ref{fig:9}~(a). All the models are trained under the $1 \times$ schedule as stated above, and the training losses of the last 8 epochs are plotted. The solid curves indicate the smoothed losses of different models, while the light-colored areas denote the standard deviations. We start from our baseline (\ie Mask R-CNN \cite{he2017mask}) and gradually incorporate the three modules to study the overall converge processes. The comparison indicates that all the proposed modules, especially DenseFPN, contribute to faster learning and higher performance, as the loss values decline faster and can reach lower minimums. Notably, the full version of CATNet converges about $30\%$ faster than the baseline while reaching a $10\%$ lower final loss.

\vspace{0.5mm}\textbf{Discriminativeness of Features:} To verify that the proposed method can effectively enhance the inadequate features and make them more discriminative, we conduct two studies on the hidden states. First, we visualize the RoI features for some common objects in iSAID dataset in Fig.~\ref{fig:1}. The feature maps are obtained by randomly select a channel from the RoI features. By aggregating context gradually from different domains, clearer object boundaries can be seen from the feature maps. Second, we randomly sampled $1,000$ objects from each category, extracted their average pooled RoI features, and plotted them on a low dimensional manifold using t-SNE \cite{van2008visualizing}. The results are shown in Fig.~\ref{fig:11}. In our baseline method, features of some categories are heavily confounded, since the top-view and low quality images cannot provide enough information for perception. By introducing multi-scale context aggregation, our method makes the features much more discriminative, as the clusters are more isolated, while features for semantically related categories (\eg \textit{large vehicle} \& \textit{small vehicle} and \textit{helicopter} \& \textit{plane}) remain close. We also compared the averaged class-wise variances of features, proving that the enhanced feature clusters are more cohesive, thus more discriminative.

\subsection{Detailed Comparisons and Ablation Study}

To study the effectiveness of the proposed modules individually, we conduct extensive experiments to compare different settings with some representative methods. All the experiments are performed on iSAID val set using the standard $1 \times$ training recipe with $512 \times 512$ patches inputs. To verify the lightweight characteristics of our method, which means the modules merely introduce negligible extra parameters and computations to the baselines, \#Params and FLOPs are reported to compare the space and time complexities of models.

\vspace{0.5mm}\textbf{Plug-and-Play Abilities:} We first verify the plug-and-play abilities of the proposed modules by incorporating them into different instance segmentation models. Aside from Mask R-CNN, we choose Cascade R-CNN \cite{cai2018cascade}, Mask Scoring R-CNN \cite{huang2019mask}, and PointRend \cite{kirillov2019pointrend} as base models, and sequentially incorporate DenseFPN, SCP, and HRoIE into these frameworks. The class-wise instance segmentation performances are reported in Table~\ref{tab:6}. We observe that for both object detection and instance segmentation tasks, our methods can steadily boost performances. Among the three proposed modules, DenseFPN provides the most significant improvement on mAPs, while SCP and HRoIE also bring considerable gains with few extra parameters. We claim that the proposed modules shall be compatible with any CNN-based instance segmentation pipelines.

\vspace{0.5mm}\textbf{Types of Multi-Scale Feature Propagation Modules:} We then study the accuracies and efficiencies of different multi-scale feature propagation modules. Table~\ref{tab:7}~(a) presents a detailed comparison among different feature pyramid networks. The results are also visualized in Fig.~\ref{fig:9} (b) and (c). Compared with existing representative methods, DenseFPN works distinctly better on both object detection and instance segmentation tasks with fewer computational costs. It is worth noting that a single-layer DenseFPN can already obtain considerable gains from the baseline with fewer or comparable computational costs. Simply stacking more basic blocks in DenseFPN can further boost the performances, indicating its capability and flexibility of model scaling.

\vspace{0.5mm}\textbf{Types of Spatial Context Modules:} Table~\ref{tab:7}~(b) shows the comparison among multiple spatial context modules. Compared with the baseline, NLNet \cite{wang2018non} can effectively bring higher performance with a high computational cost. GCNet \cite{cao2023global} solves the problem of computational complexity, but leads to an extra information confounding problem. With the help of re-weighting context, our proposed CABlock steadily outperforms GCNet using different channel reduction rates. In previous studies, these spatial attention modules are used as plugins of backbones to enhance their global context modeling abilities for image classification. However, we observe that such a strategy is sub-optimal for some modules, due to the large computational consumption with limited performance gains, as can be seen in the table. In this work, we propose to move these modules from the backbone to the layers after FPN, so that the input channels can be largely reduced, while the semantics in the feature maps are more consistent. The comparison of different designs of CABlocks (\ie used as backbone plugins or incorporated into SCP) shows that the architecture of SCP brings more performance gains with negligible computational costs.

\vspace{0.5mm}\textbf{Order of DenseFPN and SCP:} Since both DenseFPN and SCP are of pyramid structures for multi-scale features, the order of these two modules could be changed. We compare the different orders of these modules in Table~\ref{tab:7}~(c). The first row means using CABlocks are backbone plugins and adopt DenseFPN after backbone. This design is the de-facto standard for most existing works. However, it can only lead to sub-optimal performances. The second row is placing SCP before DenseFPN, so that spatial context is aggregated before feature context. This order still cannot perform well enough. The last row shows our final design, which places SCP after DenseFPN, achieving the best performance on both object detection and instance segmentation. We argue that this is because the context aggregation process shall be performed in a coarse-to-fine manner, in which we need to 1) calibrate the feature semantics among feature pyramid levels, 2) aggregate spatial context in each level, and 3) refine the instance-specific context for each object. These three steps establish the correct order of DenseFPN, SCP, and HRoIE.

\vspace{0.5mm}\textbf{Types of Region of Interest Extractors:} Table~\ref{tab:7}~(d) shows the comparison between multiple RoI extractors. The baseline model only crops the RoI features from a single feature map, leading to severe information loss and achieving ordinary results. Simply computing the sum or concatenation of RoI features cropped from multiple layers can slightly boost the performance. Considering that object detection and instance segmentation tasks require different features, incorporating HRoIE for adaptive feature fusion can better generate appropriate RoI features for these tasks.

\vspace{0.5mm}\textbf{Ablation Study and Efficiency Comparisons:} We finally conduct ablation studies to validate the effectiveness of each component. As shown in Table~\ref{tab:7}~(e), all the three proposed modules can marginally bring better results on object detection and instance segmentation in remote sensing images. When collaborating with each other, the performance improvements are still stable, indicating that these modules do not interfere with each other. The best results can be achieved by combining all these modules together, enabling aggregating multi-scale context from multiple domains simultaneously. Aside from FLOPs, we also compare the inference speed with and without the proposed modules. The FPS results are computed by averaging the model inference speed on $2,000$ images on a single A100 GPU. Compared with the baseline, adding one or two of the proposed modules leads to $\sim$$10\%$ efficiency drop, while the full version is $\sim$$20\%$ slower. Since all the modules are designed to be lightweight, the inference speeds are still acceptable when incorporating them with the baseline. Furthermore, we also observe that the higher latency mainly comes from the number of DenseFPN layers and channels in HRoIE. Therefore, a lite version of CATNet which leverages a 1-layer DenseFPN with $128$ channels is proposed. This model has a higher inference speed than the baseline, while maintaining better performances.

\section{Conclusion and Future Work}\label{sec:5}

In this work, we provided an in-depth analysis on global context modeling in remote sensing images and proposed CATNet, a novel framework that leverages three lightweight plug-and-play modules, \ie dense feature pyramid network (DenseFPN), spatial context pyramid (SCP), and hierarchical region of interest extractor (HRoIE), to aggregate the global visual context in \textit{feature}, \textit{spatial}, and \textit{instance} domains, respectively. It has been demonstrated that the collaboration among these modules can effectively enhance the discriminative object features for promoting both object detection and instance segmentation performances. We expect that the new understanding of global context and the design of proposed modules will benefit future research in this area. Below we discuss the limitations and future work in this direction.

The motivation of this work is to design lightweight plug-and-play modules for existing instance segmentation pipelines. Each proposed modules aggregate global context in one domain only. Such a strategy may be sub-optimal since cross-domain context could be correlated with each other. Therefore, a more unified framework shall be designed from scratch to mitigate the limitations of existing instance segmentation models. The recently introduced transformers \cite{vaswani2017attention} might be a better solution to handle the global context in multiple domains. We foresee the potential of incorporating multi-scale context aggregation in transformer-based models.

\IEEEpeerreviewmaketitle
\footnotesize

\bibliographystyle{IEEEtran}
\bibliography{main}

\end{document}